\pdfoutput=1

\documentclass[11pt]{article}

\usepackage[final]{acl}

\usepackage{times}
\usepackage{latexsym}

\usepackage[T1]{fontenc}

\usepackage{subcaption}
\usepackage{amsfonts}
\usepackage{multirow}
\usepackage{color}
\usepackage{soul}
\usepackage{pifont} 

\usepackage{verbatim}     
\usepackage{caption}      
\usepackage{float}        
\floatstyle{ruled} 
\newfloat{prompt}{t}{lop}
\floatname{prompt}{Prompt}

\usepackage[utf8]{inputenc}

\usepackage{microtype}

\usepackage{inconsolata}

\usepackage{graphicx}

\usepackage{paralist}

\title{Entity-aware Cross-lingual Claim Detection for Automated Fact-checking}

\author{Rrubaa Panchendrarajan, Arkaitz Zubiaga \\
  School of Electronic Engineering and Computer Science \\
  Queen Mary University of London \\
  \texttt{\{r.panchendrarajan, a.zubiaga\}@qmul.ac.uk}}

\begin{document}
\maketitle
\begin{abstract}
Identifying claims requiring verification is a critical task in automated fact-checking, especially given the proliferation of misinformation on social media platforms. Despite notable progress, challenges remain—particularly in handling multilingual data prevalent in online discourse. Recent efforts have focused on fine-tuning pre-trained multilingual language models to address this. While these models can handle multiple languages, their ability to effectively transfer cross-lingual knowledge for detecting claims spreading on social media remains under-explored. In this paper, we introduce \textit{EX-Claim}, an entity-aware cross-lingual claim detection model that generalizes well to handle multilingual claims. The model leverages entity information derived from named entity recognition and entity linking techniques to improve the language-level performance of both seen and unseen languages during training. Extensive experiments conducted on three datasets from different social media platforms demonstrate that our proposed model stands out as an effective solution, demonstrating consistent performance gains across 27 languages and robust knowledge transfer between languages seen and unseen during training.
\end{abstract}

\section{Introduction}
Automated fact-checking is an emerging research task dedicated to combating misinformation, particularly prevalent on social media, and comprises several key stages: claim detection, claim prioritization, evidence retrieval, and claim validation \citep{zeng2021automated}. While research in the task is progressing rapidly, identifying and validating claims related to global concerns requires a fact-checking pipeline capable of processing claims written in multiple languages. However, developing multilingual solutions for fact-checking research is not straightforward due to the availability of limited training data, especially for low-resource languages \citep{panchendrarajan2024claim}. This necessitates advancements in multilingual fact-checking research to leverage limited multilingual training data for developing generalized solutions that can effectively transfer knowledge between languages. 

This research focuses on the first component of the fact-checking pipeline: verifiable claim detection. A \textit{verifiable} claim is defined as a statement expressing facts, excluding personal experience and private knowledge \citep{panchendrarajan2024claim}. Existing works on verifiable claims detection are predominantly focused on monolingual solutions \citep{prabhakar2020claim,suri2022asatya,henia2021icompass,hussein2021damascusteam}, with limited attention given to addressing the multilingual nature of the problem. Especially, fine-tuning multilingual language models such as mBERT \citep{alam2020fighting,uyangodage2021can,panda2021detecting} and XLM-R \citep{alam2020fighting,husunbeyi2022rub,savchev2022ai,alam2021fighting} is observed as a common solution. While these models can handle multiple languages, none of them adequately validate the cross-lingual knowledge-transferring capability of these models, particularly in identifying claims written in languages that were not seen during training. To address this challenge, we propose \textit{EX-Claim}, a cross-lingual claim detection model that can effectively identify verifiable claims regardless of their language. Our model broadens the ability to perform across a diverse set of languages beyond those seen in training by incorporating entity-centric information for knowledge transfer. Our approach leverages the limited multilingual data available in claim detection research to develop a robust cross-lingual model that generalizes beyond the languages seen in the training phase.

We build our model on the assumption that factual claims circulating in social media often revolve around entities. In particular, entity types and their statuses such as their popularity may affect the \textit{verifiability} of a claim. For instance, consider the following three statements. 
\begin{compactitem}
    \item \textit{S1: \textbf{X} visited \textbf{China} to attend \textbf{EMNLP}}
    \item \textit{S2: \textbf{Keir Starmer} visited \textbf{France} to attend \textbf{Euro Con}}
    \item \textit{S3: \textbf{Keir Starmer} visited his home in the \textbf{UK}}
\end{compactitem}
Statement $S1$ can be deemed unverifiable from the perspective of a fact-checker, as it describes a personal experience of an unknown individual \textit{X}. However, a similar context of attending an event expressed in $S2$ is verifiable as the popular entity \textit{Keir Starmer} (the UK Prime Minister at the time of writing) attending an official event \textit{Euro Con} becomes verifiable. However, changing the popular entities \textit{Keir Starmer} or \textit{Euro Con} with another popular \textit{person} or \textit{event} entity will not change the verifiability status of this claim. At the same time, changing the \textit{event} with a private location such as \textit{home} reduces the verifiability of the claim $S3$ even if a popular entity is involved. This motivates our intuition that entities, their types, their popularity, and the relationship between them affect the verifiability status of the claims.  

Our proposed \textit{EX-Claim} builds on this intuition by leveraging entity information to effectively identify verifiable claims in a cross-lingual environment. Our model incorporates entity information such as type and popularity derived from named entity recognition (NER) and entity linking (EL). Extensive experiments conducted using three datasets from different social media platforms comprising 27 languages show that \textit{EX-Claim} stands out as the effective model with consistent performance gain across multiple languages and robust knowledge transfer between languages seen and unseen in training. We make the following key contributions:
\begin{compactitem}
    \item We propose \textit{EX-Claim}, a cross-lingual claim detection model for identifying verifiable claims from social media which can effectively transfer knowledge across languages seen and unseen during training.
    \item We enhance language-level performance by leveraging entity information in claims through NER and EL.
    \item We conduct extensive experiments to evaluate \textit{EX-Claim} across three datasets representing different social media platforms and encompassing 27 languages, including synthetic data created via machine translation.
\end{compactitem}

The source code and the synthetic data are available online\footnote{https://github.com/RubaP/Ex-Claim}.

\section{Related Work}

A common approach to identifying multilingual claims shared on social media involves fine-tuning pre-trained multilingual language models such as mBERT \citep{alam2020fighting,uyangodage2021can,panda2021detecting,zengin2021tobb,hasanain2022cross} and XLM-R \citep{alam2020fighting,alam2021fighting} and developing language-specific models to achieve optimal performance \citet{husunbeyi2022rub,savchev2022ai,eyuboglu2023fight}. Most of these works \citep{alam2020fighting,alam2021fighting,uyangodage2021can,schlicht2021upv,du2022nus} fine-tune the language model using combined training data of multiple languages and evaluate the capability of the model in identifying claims written in the same set of languages. Very few studies \citep{panda2021detecting,zengin2021tobb,hasanain2022cross} explore the model's ability to transfer knowledge across language pairs, and these studies are limited to a narrow selection of languages. 

\begin{figure*}[t]
\centering    
\includegraphics[width=0.78\textwidth]{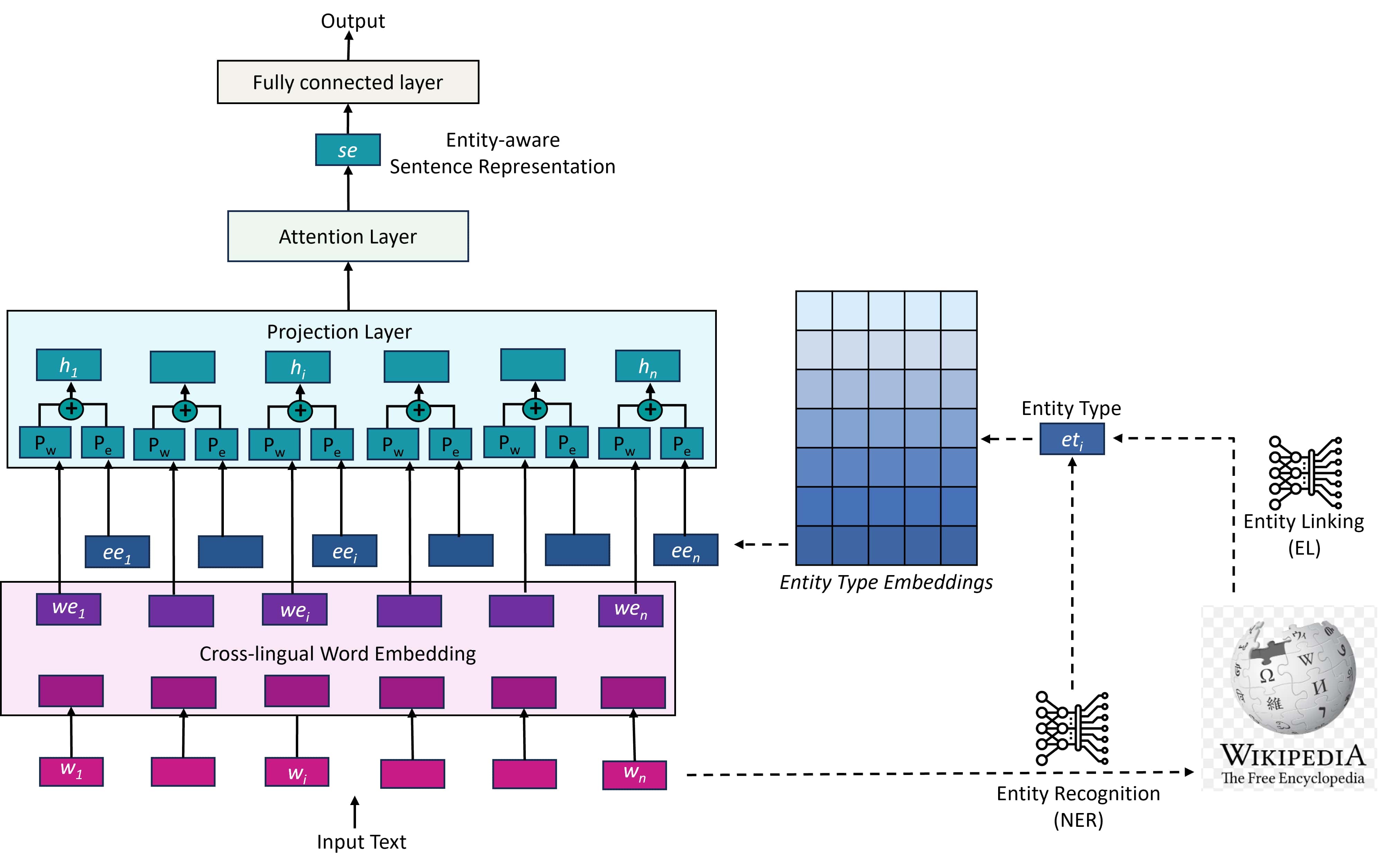}
\caption{\textit{EX-Claim :} Entity-aware Cross-lingual Claim Detection}
\label{fig:verifiable_cd_model}
\end{figure*}

With the recent attention to large language models (LLM), \citet{agrestia2022polimi} utilized GPT-3 to fine-tune the model in each language to develop a language-specific claim detection model. Interestingly, the results revealed that transformer-based architectures like BERT remain competitive with, and in some cases outperform the large language models (LLMs). Following these findings, further research \citep{sawinski2023openfact,lewoniewski2024openfact,li2024factfinders} explored leveraging LLMs for claim detection by modeling the task as a text generation problem. While these models demonstrate superior performance in the English language, their ability to handle multilingual claims for the claim detection task has yet to be explored.

Beyond directly using multilingual language models, various methods have been applied to tackle limited training data and noisy social media text. This includes the extraction of platform-specific textual features \citep{alam2020fighting}, data augmentation techniques such as machine translation \citep{zengin2021tobb,savchev2022ai} and sampling \citep{zengin2021tobb} and multitasking \cite{schlicht2021upv,du2022nus}. While these techniques are explored in multilingual settings, none of them have been shown to improve cross-lingual knowledge-transferring capabilities, which our study performs comprehensively for the first time looking at 27 different languages through the exploration of entity-based knowledge transfer. 

\section{Methodology}
Given a statement written in any language, we classify it as verifiable or non-verifiable. Figure \ref{fig:verifiable_cd_model} illustrates the proposed solution, \textit{EX-Claim}: entity-aware cross-lingual claim detection framework, whose components are discussed next. 

\subsection{Cross-lingual Word Embedding}
We use XLM-R \cite{xlm-r} to extract the cross-lingual vector representation of the words. XLM-R was trained on CommonCrawl data supporting 100 languages, and generates embeddings of size 768 for each tokenized word.  

\subsection{Named Entity Recognition (NER)}\label{subsec:entity_recognition}
Identifying entities mentioned in the text is essential to extract further information such as its type and popularity. We use state-of-the-art multilingual NER model MultiNERD \cite{tedeschi-navigli-2022-multinerd} for identifying named entities. As the model wasn't released by the authors, we fine-tuned the MultiNERD model with recommended parameters using their source code and training data. MultiNERD recognizes 15 types of fine-grained named entities.  Refer to Appendix \ref{appendix:NER_stats} for the named entity types and NER statistics.  
  
\subsection{Entity Linking (EL)}\label{subsec:entity_linking}
We consider an entity \textit{popular} if it can be linked to a Wikipedia article. Hence, we run EL for all entities identified in the previous stage to determine if an entity has an associated Wikipedia article that can be linked to.  
To do this, we make use of a state-of-the-art multilingual EL model, mGENERE \cite{decao2020multilingual} for linking entities. Given an entity-tagged sentence,
mGENERE outputs a ranked list of $n$ Wikipedia page titles along with their associated log probabilities, from which we make use of the top item with the highest probability to determine if indeed the entity has a relevant Wikipedia article that can be linked to.
If the log probability is above a threshold, we consider that entity as popular. 

During parameter tuning, we observed that the variation in threshold did not significantly affect the performance of the claim detection models that utilize EL. This is likely because most named entities were already recognized as Wikipedia entities, making the additional information redundant—possibly due to the NER model’s training data being partially sourced from Wikipedia. Consequently, we set the EL threshold to -0.15 (probability  0.861), which notably reduced the number of linked entities (refer to Figure \ref{fig:WikiNER_Link_stat} in Appendix \ref{appendix:entity_linking_stats}), ensuring that only highly probable entity links are used to determine the popularity. Refer to Appendix \ref{appendix:entity_linking_stats} for the EL statistics.   

\subsection{Entity Type Embedding}
Similar to the word embedding, an entity type can be represented in a dense vector representation, enabling the model to capture entity type information required for the task as a dense vector. This is achieved by adding an embedding layer to the claim detection model, which converts each entity type to an entity type vector. Based on the entity type leveraged, we experiment with the following two variants of \textit{EX-Claim}:
\begin{compactitem}
    \item \textit{EXN-Claim}: Uses 15 named entity types recognized by the NER model, along with the \textit{other} category for non-entities—resulting in a total of 16 entity types.
    \item \textit{EXP-Claim}: Uses both NER and the popularity of an entity. Each named entity can be either popular or unpopular,  doubling the size of the entity types. This results in 31 entity types (2*15 + 1).
\end{compactitem}

As an ablated variant, we experiment with \textit{X-Claim} category which does not use any entity-type information to show the effectiveness of utilizing entity-type information for verifiable claim detection. Except for the \textit{X-Claim} category, each word in the input sequence is identified with an entity type index $et_i$ according to one of the entity type categories. This information is fed into the entity type embedding layer to obtain a dense vector representation $ee_i$ for each word in the input sequence.

\subsection{Entity-aware Claim Detection}
\begin{table}[t]
\footnotesize
\centering
\label{tab:notations}
\begin{tabular}{lp{5cm}}\hline
Notation & Description \\ \hline
$w_i$         & $i^{th}$ word from the input sequence             \\
$n$         & Number of words in the input sequence            \\
$we_i \in \mathbb{R}^{d_w}$         & Word embedding of $i^{th}$ word            \\
$d_w$         &  Dimension of word embedding            \\ 
$et_i \in \mathbb{R}^k$         & One-hot vector indicating entity type of $i^{th}$ word            \\
$k$         &  Number of entity types            \\
$ee_j \in \mathbb{R}^{d_e}$         & Entity type embedding of $j^{th}$ entity type            \\
$d_e$         &  Dimension of entity type embedding            \\ \hline
$d_p$         &  Projection layer dimension            \\
$P_w \in \mathbb{R}^{d_w \times d_p}$         &  Word embedding projection matrix            \\
$P_e \in \mathbb{R}^{d_e \times d_p}$         &  Entity type embedding projection matrix            \\
$h_i \in  \mathbb{R}^{d_p} $ & Projected input representation \\ 
$H$ & Projected input sequence $(h_1, h_2, ...h_n)$  \\ \hline
$se \in  \mathbb{R}^{d_p}$ & Sentence embedding \\ \hline
$C$ & Number of class labels \\
$y \in \mathbb{R}^{C}$ & Probability distribution predicted by the model for the input sequence\\
$\hat{y} \in \mathbb{R}^{C}$ & Ground truth probability distribution of the input sequence \\ \hline
$EE \in \mathbb{R}^{k \times d_e}$         &  Entity embedding layer            \\
$Attn$ & Attention layer \\
\hline
\end{tabular}
\caption{List of Key Notations}
\end{table}

Given an input word sequence $(w_1, w_2, \ldots, w_n)$, we extract the cross-lingual word embeddings $(we_1, we_2, \ldots, we_n)$ , and the entity type index $(et_1, et_2, \ldots, et_n)$. Here, $et_i \in \mathbb{R}^k$ represents the entity type of a word as a one-hot vector, where $k$ indicates the number of entity types. These entity-type indices are used to extract entity-type embeddings as follows,
\begin{equation}
\small
    EE(et_1, et_2,... et_n) = (ee_1, ee_2,... ee_n)
\end{equation}

Once the entity-type embeddings are obtained, we project both word embeddings and entity-type embeddings into a common vector space and add the projected vectors at the word level as follows,
\begin{equation}
\small
    h_i = P_w \cdot we_i + P_e \cdot ee_i \in \mathbb{R}^{d_p}
\end{equation}

This projection layer is responsible for converting the input features into task-specific, low-dimensional vectors and incorporating entity-type information into the word embedding. 

To attend to the important words indicating the verifiability of claims, we apply scaled dot-product attention \citep{vaswani2017attention} to the projected input sequence $H = (h_1, h_2, ...h_n)$ as follows,
\begin{equation}
\small
    Attn(H) = softmax(\frac{HH^T}{\sqrt{d_p}}) \cdot H
\end{equation}

The attention layer $Attn$ generates attended context vectors for each element in the input sequence. The mean value of this context vector is obtained as the sentence representation $se$ as follows,
\begin{equation}
\small
    se = \frac{1}{n}\sum Attn(H) \in \mathbb{R}^{2d_p}
\end{equation}

Finally, the sentence embedding $se$ is fed into a fully connected layer followed by a softmax function to generate a probability distribution $y$ over the output classes as follows,
\begin{equation}
\small
    out = W_o se + b_o \in \mathbb{R}^{C}
\end{equation}
\begin{equation}
\small
    y_j = \frac{exp(out_j)}{\sum_j^C exp(out_j)} \in \mathbb{R}^{C}
\end{equation}

Given the true label distribution $\hat{y} \in \mathbb{R}^{C}$, the entity type embedding layer, projection layer, and attention layer are trained by optimizing the cross-entropy loss. 

\section{Experiment Setting}

\subsection{Dataset}
We use CheckThat! 2022 \cite{nakov2022overview} dataset for training and testing the claim detection models. This dataset contains tweets related to COVID-19, written in 5 languages. Further, it contains four partitions: Train, Dev, Dev-test, and Test. We used the Train partition of all languages except Arabic for training and used the Dev partition as validation data. The Arabic data was excluded during the training as the initial experiments yielded poor performance in all languages including Arabic. 
Further, the Dev-Test partition was used to find the optimal parameters.

To evaluate the models’ ability to identify claims written in a broad range of unseen languages, we generated synthetic data from the CheckThat! Test partition for 18 unseen languages. For each, we randomly sampled 250 instances and translated them using Microsoft Azure AI translator \cite{junczys2019microsoft}, limiting samples to match the test dataset size. Translation quality was assessed on the FLORES-200 benchmark \cite{nllb2022}, with a CHRF score of 54.35. Further details are in Appendix \ref{sec:machine_translation_performance}.

To evaluate the generalization capability of the proposed model in identifying claims discussing unseen topics as well as unseen language, we report the performance in Kazemi 2021 \cite{kazemi2021claim}. This dataset contains WhatsApp tipline and social group messages written in 5 languages. Refer to Appendix \ref{sec:dataset_stats} for the statistics of the datasets. 

\begin{table*}[t]
\footnotesize
\centering
\begin{tabular}{lllll} \hline
                  Model     & Accuracy    & Precision   & Recall      & F1-Score    \\ \hline 
                  \multicolumn{5}{c}{CheckThat!   2022}                             \\ \hline
                  mBERT     & 0.722 $\pm$ 0.006 & 0.723 $\pm$ 0.006 & 0.722 $\pm$ 0.006 & 0.715 $\pm$ 0.007 \\
                  XLM-R      & 0.669 $\pm$ 0.007 & 0.695 $\pm$ 0.006 & 0.669 $\pm$ 0.007 & 0.658 $\pm$ 0.009 \\
                  mT5      & 0.681 $\pm$ 0.004 & 0.685 $\pm$ 0.004 & 0.681 $\pm$ 0.004 & 0.679 $\pm$ 0.004 \\\hline
                  Llama2 & \textbf{0.776} $\pm$ 0.001  & 0.782 $\pm$ 0.001  & \textbf{0.776} $\pm$ 0.001  & \textbf{0.771} $\pm$ 0.001  \\
                  Mistral & 0.769 $\pm$ 0.003  & \textbf{0.784} $\pm$ 0.003  & 0.769 $\pm$ 0.003  & 0.761 $\pm$ 0.003  \\
                  Phi3 & 0.744 $\pm$ 0.001  & 0.75 $\pm$ 0.001  & 0.744 $\pm$ 0.001  & 0.738 $\pm$ 0.001  \\\hline
X-Claim      & 0.747 $\pm$ 0.005 & 0.751 $\pm$ 0.005 & 0.747 $\pm$ 0.005 & 0.745 $\pm$ 0.005 \\
                  EXN-Claim      & 0.754 $\pm$ 0.005 & 0.758 $\pm$ 0.006 & 0.754 $\pm$ 0.005 & 0.752 $\pm$ 0.006 \\
                  EXP-Claim & 0.755 $\pm$ 0.007 & 0.759 $\pm$ 0.007 & 0.755 $\pm$ 0.007 & 0.754 $\pm$ 0.007 \\ \hline
                  \multicolumn{5}{c}{Synthetic   Data}                              \\ \hline
                  mBERT     & 0.687 $\pm$ 0.009 & 0.704 $\pm$ 0.008 & 0.687 $\pm$ 0.009 & 0.679 $\pm$ 0.01  \\
                  XLM-R      & 0.662 $\pm$ 0.008 & 0.68 $\pm$ 0.007  & 0.662 $\pm$ 0.008 & 0.656 $\pm$ 0.009 \\ 
                  mT5      & 0.659 $\pm$ 0.007 & 0.662 $\pm$ 0.007 & 0.659 $\pm$ 0.007 & 0.658 $\pm$ 0.007 \\\hline
                  Llama2 & 0.724 $\pm$ 0.002  & 0.749 $\pm$ 0.002  & 0.724 $\pm$ 0.001  & 0.707 $\pm$ 0.002  \\
                  Mistral & 0.718 $\pm$ 0.002  & \textbf{0.759} $\pm$ 0.003  & 0.718 $\pm$ 0.002  & 0.701 $\pm$ 0.003  \\
                  Phi3 & 0.691 $\pm$ 0.002  & 0.735 $\pm$ 0.005  & 0.691 $\pm$ 0.002  & 0.664 $\pm$ 0.002  \\\hline
X-Claim      & 0.719 $\pm$ 0.007 & 0.724 $\pm$ 0.006 & 0.719 $\pm$ 0.007 & 0.718 $\pm$ 0.007 \\
                  EXN-Claim      & 0.73 $\pm$ 0.008  & 0.735 $\pm$ 0.007 & 0.73 $\pm$ 0.008  & 0.729 $\pm$ 0.008 \\
                  EXP-Claim & \textbf{0.731} $\pm$ 0.009 & 0.737 $\pm$ 0.008 & \textbf{0.731} $\pm$ 0.009 & \textbf{0.73} $\pm$ 0.009  \\ \hline
                  \multicolumn{5}{c}{Kazemi   2021 Data}                            \\ \hline
                  mBERT     & 0.716 $\pm$ 0.007 & 0.725 $\pm$ 0.006 & 0.716 $\pm$ 0.007 & 0.702 $\pm$ 0.009 \\
                  XLM-R      & 0.687 $\pm$ 0.011 & 0.752 $\pm$ 0.004 & 0.687 $\pm$ 0.011 & 0.686 $\pm$ 0.012 \\ 
                  mT5      & 0.681 $\pm$ 0.004 & 0.725 $\pm$ 0.004 & 0.681 $\pm$ 0.004 & 0.678 $\pm$ 0.005 \\\hline
                  Llama2 & 0.633 $\pm$ 0.003  & 0.5 $\pm$ 0.001  & 0.633 $\pm$ 0.001  & 0.613 $\pm$ 0.004  \\
                  Mistral & 0.603 $\pm$ 0.004  & 0.747 $\pm$ 0.005  & 0.603 $\pm$ 0.004  & 0.55 $\pm$ 0.005  \\
                  Phi3 & 0.546 $\pm$ 0.001  & 0.744 $\pm$ 0.001  & 0.546 $\pm$ 0.001  & 0.455 $\pm$ 0.001  \\\hline
X-Claim      & 0.788 $\pm$ 0.007 & \textbf{0.817} $\pm$ 0.005 & 0.788 $\pm$ 0.007 & 0.789 $\pm$ 0.007 \\
                  EXN-Claim      & 0.793 $\pm$ 0.008 & 0.813 $\pm$ 0.006 & 0.793 $\pm$ 0.008 & 0.796 $\pm$ 0.008 \\
                  EXP-Claim & \textbf{0.794} $\pm$ 0.008 & 0.813 $\pm$ 0.006 & \textbf{0.794} $\pm$ 0.008 & \textbf{0.797} $\pm$ 0.007 \\ \hline
                  \multicolumn{5}{c}{All}                                           \\ \hline
                  mBERT     & 0.698 $\pm$ 0.008 & 0.711 $\pm$ 0.007 & 0.698 $\pm$ 0.008 & 0.689 $\pm$ 0.009 \\
                  XLMR      & 0.668 $\pm$ 0.008 & 0.695 $\pm$ 0.006 & 0.668 $\pm$ 0.008 & 0.661 $\pm$ 0.01  \\ 
                  mT5      & 0.667 $\pm$ 0.006 & 0.677 $\pm$ 0.006 & 0.667 $\pm$ 0.006 & 0.665 $\pm$ 0.006 \\\hline
                  Llama2 & 0.717 $\pm$ 0.002  & 0.755 $\pm$ 0.001  & 0.717 $\pm$ 0.002  & 0.702 $\pm$ 0.002  \\
                  Mistral & 0.707 $\pm$ 0.003  & 0.761 $\pm$ 0.004  & 0.707 $\pm$ 0.003  & 0.685 $\pm$ 0.003  \\
                  Phi3 & 0.674 $\pm$ 0.002  & 0.74 $\pm$ 0.004  & 0.674 $\pm$ 0.002  & 0.64 $\pm$ 0.002  \\\hline
X-Claim      & 0.736 $\pm$ 0.006 & 0.746 $\pm$ 0.006 & 0.736 $\pm$ 0.006 & 0.735 $\pm$ 0.007 \\
                  EXN-Claim      & 0.746 $\pm$ 0.007 & 0.753 $\pm$ 0.007 & 0.746 $\pm$ 0.007 & 0.745 $\pm$ 0.008 \\
                  EXP-Claim & \textbf{0.747} $\pm$ 0.008 & \textbf{0.754} $\pm$ 0.007 & \textbf{0.747} $\pm$ 0.008 & \textbf{0.746} $\pm$ 0.009 \\ \hline
\end{tabular}
\caption{Dataset-level Performance of the Claim Detection Models}\label{tab:performance_datasets}
\end{table*}

\subsection{Evaluation metrics}
We report accuracy, precision, recall, and F1-score for ten fine-tuned claim detection models, presenting their averages and standard deviations. To account for varying ratios of verifiable and unverifiable claims across languages, we calculated weighted scores at the language level, and the overall performance is obtained by averaging the language-level performance of the models.

\subsection{Baseline models}
Due to the lack of cross-lingual claim detection models, we use strong cross-lingual language models as baselines: mBERT \citep{alam2020fighting,uyangodage2021can,panda2021detecting,zengin2021tobb,hasanain2022cross}, XLM-RoBERTa (XLM-R) \citep{alam2020fighting,alam2021fighting}, and mT5 \citep{du2022nus}, all fine-tuned on the CheckThat! Train partition. These widely used models have proven competitive in multilingual claim detection. We also include open-source large language models from \citet{li2024factfinders}—Llama2-7B, Mistral-7B, and Phi3—fine-tuned on the same data for text generation.

Refer to Appendix \ref{sec:implementation} for resources used for the training, hyperparameters, the prompt used for LLMs, and the LLM fine-tuning process. 

\section{Results} 

\subsection{Dataset-level Performance}

Table \ref{tab:performance_datasets} summarizes the performance of the claim detection models. Llama2 performs strongly on the CheckThat! 2022 Test dataset, while \textit{EXP-Claim} achieves the highest accuracy, recall, and F1-score on both the synthetic and Kazemi datasets. The performance of large language models declines on these two datasets, highlighting challenges in handling linguistically diverse data.

Notably, the EX-Claim models and their ablated variants outperform both transformer-based baselines and large language models in overall evaluation, underscoring the value of learning task-specific representations that focus on claim-relevant words. Moreover, their consistent performance across diverse datasets—spanning multiple platforms and topics—suggests that our hypothesis holds generally, regardless of the claim's nature.

Incorporating NER information (\textit{EXN-Claim} vs. \textit{X-Claim}) consistently improves performance across all four metrics. While the overall gain is modest (approximately 1\%), our language-level analysis (Section \ref{sec:gain_analysis}) shows that NER integration either maintains or enhances model performance across most of the languages, with certain languages benefiting significantly from entity information and linguistic knowledge transfer from the NER and EL tools. Interestingly, incorporating entity popularity (\textit{EXN-Claim} vs. \textit{EXP-Claim}) did not lead to further performance improvement—likely because most entities identified by the NER model were already recognized as popular, limiting the potential for additional gains.

\subsection{Language-level Gain}\label{sec:gain_analysis}

\begin{figure*}[t]
    \centering
    \begin{subfigure}[b]{0.79\textwidth}
    \includegraphics[width=\textwidth]{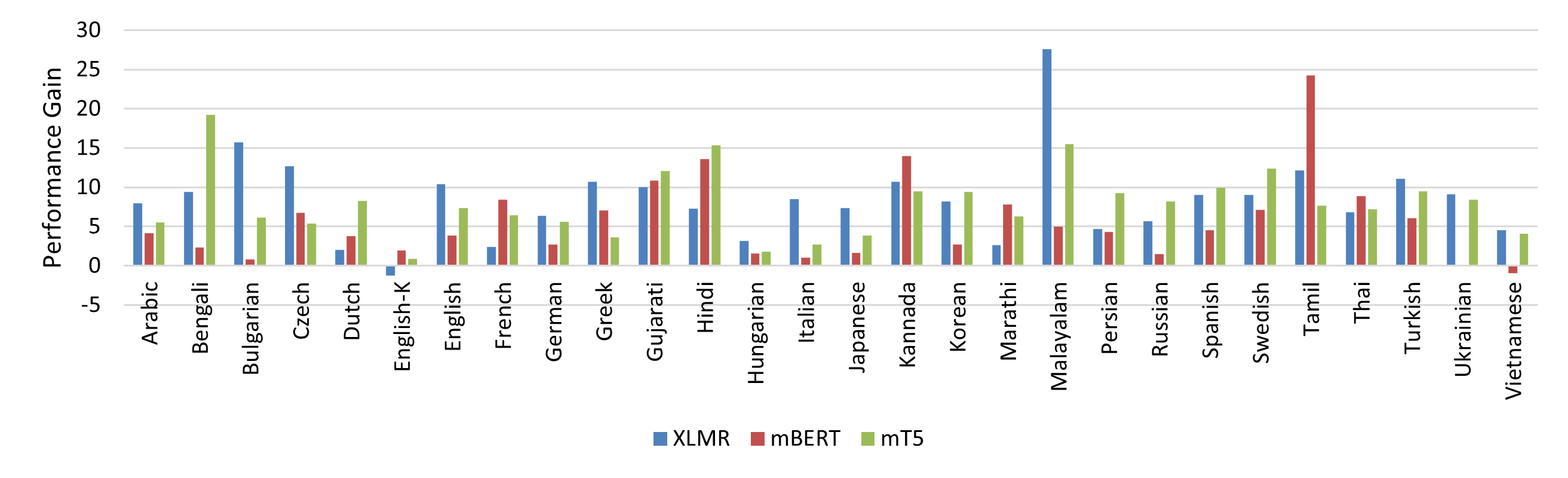}
    \caption{Transformer-based Baselines}\label{fig:gain_transformers}
    \end{subfigure}  
    \begin{subfigure}[b]{0.79\textwidth}
    \includegraphics[width=\textwidth]{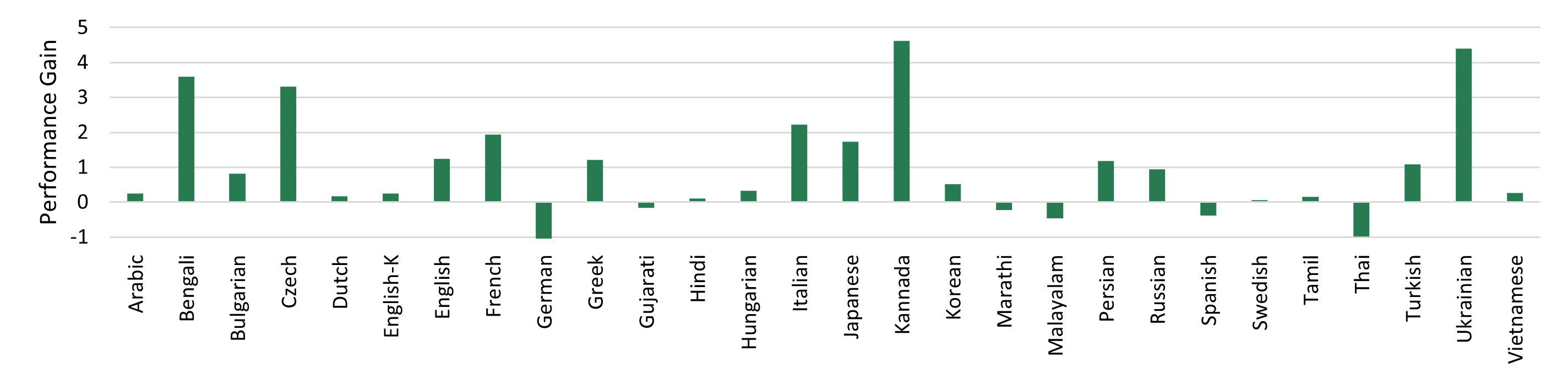}
    \caption{\textit{X-Claim}}\label{fig:gain_xclaim}  
    \end{subfigure}
    \begin{subfigure}[b]{0.79\textwidth}
    \includegraphics[width=\textwidth]{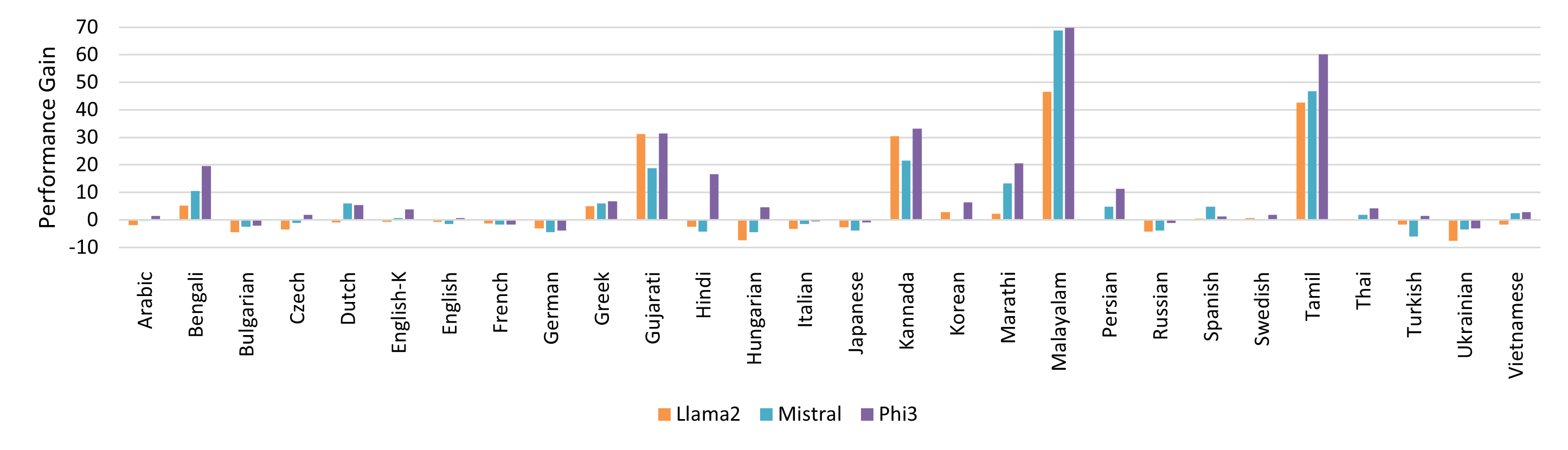}
    \caption{Large Language Models (LLMs)}\label{fig:gain_llm}  
    \end{subfigure}  
    \caption{Performance Gain in \textit{EXN-Claim} compared to Three groups of Models}
    \label{fig:performance_gain}
\end{figure*}

We analyze language-level performance gains from incorporating entity information into claim detection. Since there is no performance gain in \textit{EXP-Claim} compared to \textit{EXN-Claim}, 
we analyze the performance of \textit{EXN-Claim} with three groups of models as shown in Figure \ref{fig:performance_gain}. Refer to Appendix \ref{sec:laguage_level_performance} for the language-level F1-score of the models.

Figure~\ref{fig:gain_transformers} illustrates the performance gain of the \textit{EXN-Claim} model over transformer-based baselines. Most languages exhibit consistent positive gains, except English (Kazemi dataset) and Vietnamese, which show slight drops (0.95–1.2\%). The most substantial improvements are seen in Asian languages like Malayalam, Tamil, Hindi, Bengali, Kannada, and Gujarati. These results underscore the effectiveness of incorporating entity information and dynamically attending to claim-relevant words, leading to significant performance gains across a diverse set of languages.

Figure~\ref{fig:gain_xclaim} shows the performance gain of the \textit{EXN-Claim} model over its ablation variant, \textit{X-Claim}, which excludes entity information. Interestingly, incorporating entity information either retains or improves model performance across most languages. The only exceptions are German and Thai, which exhibit a slight performance drop of approximately 0.97\% to 1.1\%. Notably, certain languages—such as Kannada, Ukrainian, Bengali, Czech, Italian, and French—experience the highest gains, ranging from 1.95\% to 4.6\%. These improvements can likely be attributed to the incorporation of entity information and the indirect linguistic knowledge transfer from the NER tool.

Similarly, we compare the performance gain of the \textit{EXN-Claim} model with large language models (LLMs) in Figure~\ref{fig:gain_llm}. While LLMs outperform \textit{EXN-Claim} in several high-resource languages, they perform notably poorly in certain low-resource languages—particularly Asian regional languages such as Malayalam, Kannada, Tamil, Gujarati, Bengali, and Hindi. The largest gains are in Malayalam (up to 70\%) and Tamil (up to 60\%), which include non-synthetic samples from the Kazemi dataset. Despite being smaller and more computationally efficient, \textit{EXN-Claim} outperforms all three LLMs in 7 languages and surpasses at least one LLM in 16 out of the 27 languages evaluated. 

Furthermore, we observe inconsistent gains across LLMs: Llama excels in Dutch, Phi3 in Hungarian, while Mistral underperforms in both. Similarly, Llama and Phi3 perform well in Korean, but Mistral struggles. These observations suggest that LLMs tend to excel only in specific languages, largely influenced by the composition of their training data. In contrast, these findings highlight \textit{EXN-Claim} as a highly efficient model that consistently delivers superior performance across many low-resource languages, while remaining competitive with LLMs in high-resource scenarios, offering the best balance between overall performance gains and computational efficiency. Refer to Appendix \ref{sec:model_comaprison} for comparison of claim detection models based on the number of parameters, multilingual support, and training time.

\section{Analysis}

\subsection{Cross-lingual Knowledge Transfer}
The synthetic data comprises claims from the CheckThat! Test partition, written in 18 different languages. We use this dataset to assess the cross-lingual knowledge-transferring capabilities of the claim detection models. We compute the fraction of data instances in the Synthetic data which gets the same class prediction as its original data instance from the CheckThat! Test partition by a model. We categorized these results into correct and incorrect predictions, allowing us to evaluate the model's ability to transfer knowledge across correct and wrong predictions.
 
\begin{table}[t]
\footnotesize
\centering
\begin{tabular}{lll}\hline
                  Model     & Correct Prediction & Wrong Prediction \\ \hline
                  mBERT     & 82.75\%                                                        & 64.83\%                                                      \\
                  XLMR      & 81.72\%                                                        & 66.27\%                                                      \\
                  mT5      & 79.12\%                                                        & \textbf{60}\%                                                      \\\hline
                  Llama2     & 85.4\%                                                        & 70.1\%                                                      \\
                  Mistral      & 86.87\%                                                        & 71.3\%                                                      \\
                  Phi3      & 82.46\%                                                        & 65.5\%                                                      \\\hline
X-Claim      & 87.21\%                                                        & 71.73\%   \\          
                  EXN-Claim      & \textbf{87.82}\%                                                        & 68.48\%                                                      \\
                  EXP-Claim & 87.09\%                                                        & 66.51\%    \\ \hline                                                     
\end{tabular}
\caption{Cross-lingual Knowledge Transfer Rate}\label{tab:knowledge_transfer_rate}
\end{table}

Table \ref{tab:knowledge_transfer_rate} lists the knowledge transfer rate of claim detection models. Overall, the transformer-based baseline models exhibit the lowest transfer rate in correct and wrong predictions. In contrast, \textit{EX-Claim} and its ablated variant achieve the highest transfer rate—reaching nearly 87\%—for correct predictions. This highlights the effectiveness of learning context-aware claim representations with attention to claim-relevant words in promoting cross-lingual generalization. Furthermore, the inclusion of entity information through NER and EL (\textit{EXN-Claim} and \textit{EXP-Claim}) reduces the transfer of incorrect predictions by 3\% to 5\%. This indicates that entity-level knowledge and linguistic signals from NER and EL help prevent consistent misclassification of the same instances across different languages, thus improving model robustness.

\subsection{Attending to Important Words}
The attention layer enables the claim detection models to attend to important words in the input sequence before generating the sentence representation. We quantify this behavior of the models as the entropy of attention weights computed by the attention layer. We compute the average entropy of the test data by computing the mean value of entropy of all the input sequences.

\begin{table}[]
\footnotesize
\centering
\begin{tabular}{ll}\hline
Model     & Entropy \\ \hline
X-Claim      & 4.09 \\
EXN-Claim      & 3.519  \\
EXP-Claim & \textbf{3.322}  \\ \hline                                  \end{tabular}
\caption{Entropy of Attention Layer}\label{tab:entropy}
\end{table}

Table \ref{tab:entropy} lists the attention layer's entropy of the claim detection models. We can see that the models utilizing entity information always yield lower entropy, especially when utilizing popularity, resulting in the lowest entropy among all. This shows that these models are concentrated on a few keywords in the input sequence to determine the verifiability of a claim.\footnote{Refer to Appendix \ref{sec:appendix_attention_weights} for attention weights generated for sample sentences.}   

\subsection{Error Analysis}\label{sec:error_analysis}

\begin{table}[t]
\footnotesize
\centering
\begin{tabular}{lp{6.5cm}}\hline
 \multicolumn{2}{c}{False Positive} \\ \hline
S4      & \textit{Months after receiving the second dose of the Pfizer COVID vaccine, I continue to suffer side effects including perfect health and having to deal with morons.} \\
S5 & \textit{How true is Congress party's claim regarding shortfall in Government earnings} \\ \hline
\multicolumn{2}{c}{False Negative} \\ \hline
S6 & \textit{These are not photos of a man with an artificial heart} \\
S7 & \textit{Retirement Home NOT Raided By The FBI For Running Elderly Fight Club, NO 7 Arrests} \\ 
S8 & \textit{Gruesome Video Of A Man Hacked To Death In Bihar Falsely Shared As TMC 'Goons' Attacking A BJP Worker} \\ 
S9 & \textit{Video Showing Suicide Bid Falsely Linked To CAA} \\ \hline                                  \end{tabular}
\caption{Misclassified Sample Sentences}\label{tab:misclassification_examples}
\end{table}

We analyze the false positive and false negative predictions made by \textit{EX-Claim} models. Table \ref{tab:misclassification_examples} lists some of the misclassified examples. We noticed that the appearance of disease names ($S4$), lengthy claims, nonrecognition of personal experience from pronouns' presence ($S4$), and ambiguous fact expressions could cause false positive scenarios ($S5$).

We observed that misclassification of popular events expressing personal experience tends to be a common cause of false negatives. Refer to Sentences \textit{S6-S9} in Table \ref{tab:misclassification_examples} for sample false negative instances where the reach of the event makes it verifiable, yet the personal experience framing leads to misclassification. Further, some of these claims refer to other modalities such as images and videos in the expression of facts, making it further difficult when relying solely on textual information 
\footnote{Refer to Appendix \ref{appendix:correlation_matrix} showing the correlation between the confusion matrix and some of these features.}.

\section{Conclusion}
We present \textit{EX-Claim}, the first cross-lingual claim detection model that integrates entity information for improved performance in detecting claims across multiple languages. Extensive experiments on three datasets comprising 27 languages reveal that \textit{EX-Claim} stands out as the effective model with consistent performance gain across multiple languages compared to the baselines. Language-level performance analysis reveals that certain languages—particularly Asian regional languages—benefit significantly from the incorporation of entity information. This improvement is likely driven by the emphasis on key terms essential for claim detection and by the indirect transfer of linguistic knowledge from named entity recognition and entity linking tools. Further analysis of cross-lingual transfer rates and attention layer entropy supports this finding. Notably, \textit{EX-Claim} achieves the highest cross-lingual knowledge transfer rate across correct predictions while minimizing the wrong predictions across claims written in the same language, demonstrating its robustness in cross-lingual generalization and transfer efficiency. 
Future work will focus on generalizing entity-type embeddings beyond training contexts and address the challenges contributing to false-positive and false-negative outcomes identified in our error analysis.  

\section{Limitations}
The limitations of this work are as follows:
\begin{compactitem}
    \item Training data limited to a single topic - Our training data consists solely of COVID-19 discussions. Although our model generalizes well across different datasets, including those focused on political topics, this limitation may still restrict the model's broader generalization capabilities.
    \item Dependence on NER and entity linking tools: The proposed solution relies on existing cross-lingual NER and entity linking tools to extract entity-type information. The performance of these tools directly impacts the overall effectiveness of our model, particularly in languages where NER or entity-linking tools may perform poorly or exceptionally well.
    \item Training data limited to a single platform: Our training data is exclusively composed of tweets, which may cause the model to learn characteristics specific to this platform. While we demonstrate that our model can handle data from other platforms, such as WhatsApp tiplines in the Kazemi dataset, this limitation could still hinder the model's ability to generalize across different types of social media content.
    \item Exclusion of popular events not identified by a name: Since the system identifies and tracks the popularity of named entities, popular events not linked to any names are excluded. As previously discussed, this is a significant cause of false negative predictions. 
\end{compactitem}

\section*{Acknowledgments}
This project is funded by the European Union and UK Research and Innovation under Grant No. 101073351 as part of Marie Skłodowska-Curie Actions (MSCA Hybrid Intelligence to monitor, promote, and analyze transformations in good democracy practices).

\bibliography{references}

\begin{thebibliography}{33}
\providecommand{\natexlab}[1]{#1}

\bibitem[{Agrestia et~al.(2022)Agrestia, Hashemianb, and Carmanc}]{agrestia2022polimi}
S~Agrestia, AS~Hashemianb, and MJ~Carmanc. 2022.
\newblock Polimi-flatearthers at checkthat! 2022: Gpt-3 applied to claim detection.
\newblock \emph{Working Notes of CLEF}.

\bibitem[{Alam et~al.(2021{\natexlab{a}})Alam, Dalvi, Shaar, Durrani, Mubarak, Nikolov, Da~San~Martino, Abdelali, Sajjad, Darwish et~al.}]{alam2021fighting}
Firoj Alam, Fahim Dalvi, Shaden Shaar, Nadir Durrani, Hamdy Mubarak, Alex Nikolov, Giovanni Da~San~Martino, Ahmed Abdelali, Hassan Sajjad, Kareem Darwish, et~al. 2021{\natexlab{a}}.
\newblock Fighting the covid-19 infodemic in social media: a holistic perspective and a call to arms.
\newblock In \emph{Proceedings of the International AAAI Conference on Web and Social Media}, volume~15, pages 913--922.

\bibitem[{Alam et~al.(2021{\natexlab{b}})Alam, Shaar, Dalvi, Sajjad, Nikolov, Mubarak, Da~San~Martino, Abdelali, Durrani, Darwish et~al.}]{alam2020fighting}
Firoj Alam, Shaden Shaar, Fahim Dalvi, Hassan Sajjad, Alex Nikolov, Hamdy Mubarak, Giovanni Da~San~Martino, Ahmed Abdelali, Nadir Durrani, Kareem Darwish, et~al. 2021{\natexlab{b}}.
\newblock Fighting the covid-19 infodemic: Modeling the perspective of journalists, fact-checkers, social media platforms, policy makers, and the society.
\newblock In \emph{Findings of the Association for Computational Linguistics: EMNLP 2021}, pages 611--649.

\bibitem[{Baris-Schlicht et~al.(2021)Baris-Schlicht, Magnossao~de Paula, and Rosso}]{schlicht2021upv}
Ipek Baris-Schlicht, Angel~Felipe Magnossao~de Paula, and Paolo Rosso. 2021.
\newblock Upv at checkthat! 2021: Mitigating cultural differences for identifying multilingual check-worthy claims.
\newblock In \emph{Proceedings of the Working Notes of CLEF 2021, Conference and Labs of the Evaluation Forum, Bucharest, Romania, September 21st to 24th, 2021}, pages 465--475. CEUR.

\bibitem[{De~Cao et~al.(2022)De~Cao, Wu, Popat, Artetxe, Goyal, Plekhanov, Zettlemoyer, Cancedda, Riedel, and Petroni}]{decao2020multilingual}
Nicola De~Cao, Ledell Wu, Kashyap Popat, Mikel Artetxe, Naman Goyal, Mikhail Plekhanov, Luke Zettlemoyer, Nicola Cancedda, Sebastian Riedel, and Fabio Petroni. 2022.
\newblock \href {https://doi.org/10.1162/tacl_a_00460} {{Multilingual Autoregressive Entity Linking}}.
\newblock \emph{Transactions of the Association for Computational Linguistics}, 10:274--290.

\bibitem[{Devlin et~al.(2019)Devlin, Chang, Lee, and Toutanova}]{bert}
Jacob Devlin, Ming-Wei Chang, Kenton Lee, and Kristina Toutanova. 2019.
\newblock \href {https://doi.org/10.18653/v1/N19-1423} {{BERT}: Pre-training of deep bidirectional transformers for language understanding}.
\newblock In \emph{Proceedings of the 2019 Conference of the North {A}merican Chapter of the Association for Computational Linguistics: Human Language Technologies, Volume 1 (Long and Short Papers)}, pages 4171--4186, Minneapolis, Minnesota. Association for Computational Linguistics.

\bibitem[{Du et~al.(2022)Du, Gollapalli, and Ng}]{du2022nus}
SM~Du, Sujatha~Das Gollapalli, and See-Kiong Ng. 2022.
\newblock Nus-ids at checkthat! 2022: identifying check-worthiness of tweets using checkthat5.
\newblock \emph{Working Notes of CLEF}.

\bibitem[{Eyuboglu et~al.(2023)Eyuboglu, Altun, Arslan, Sonmezer, and Kutlu}]{eyuboglu2023fight}
Ahmet~Bahadir Eyuboglu, Bahadir Altun, Mustafa~Bora Arslan, Ekrem Sonmezer, and Mucahid Kutlu. 2023.
\newblock Fight against misinformation on social media: Detecting attention-worthy and harmful tweets and verifiable and check-worthy claims.
\newblock In \emph{International Conference of the Cross-Language Evaluation Forum for European Languages}, pages 161--173. Springer.

\bibitem[{Hasanain and Elsayed(2022)}]{hasanain2022cross}
Maram Hasanain and Tamer Elsayed. 2022.
\newblock Cross-lingual transfer learning for check-worthy claim identification over twitter.
\newblock \emph{arXiv preprint arXiv:2211.05087}.

\bibitem[{Henia et~al.(2021)Henia, Rjab, Haddad, and Fourati}]{henia2021icompass}
Wassim Henia, Oumayma Rjab, Hatem Haddad, and Chayma Fourati. 2021.
\newblock icompass at nlp4if-2021--fighting the covid-19 infodemic.
\newblock In \emph{Proceedings of the Fourth Workshop on NLP for Internet Freedom: Censorship, Disinformation, and Propaganda}, pages 115--118.

\bibitem[{Hussein et~al.(2021)Hussein, Ghneim, and Joukhadar}]{hussein2021damascusteam}
Ahmad Hussein, Nada Ghneim, and Ammar Joukhadar. 2021.
\newblock Damascusteam at nlp4if2021: Fighting the arabic covid-19 infodemic on twitter using arabert.
\newblock In \emph{Proceedings of the Fourth Workshop on NLP for Internet Freedom: Censorship, Disinformation, and Propaganda}, pages 93--98.

\bibitem[{H{\"u}s{\"u}nbeyi et~al.(2022)H{\"u}s{\"u}nbeyi, Deck, and Scheffler}]{husunbeyi2022rub}
Zehra~Melce H{\"u}s{\"u}nbeyi, Oliver Deck, and Tatjana Scheffler. 2022.
\newblock Rub-dfl at checkthat! 2022: Transformer models and linguistic features for identifying relevant.
\newblock \emph{Working Notes of CLEF}.

\bibitem[{Junczys-Dowmunt(2019)}]{junczys2019microsoft}
Marcin Junczys-Dowmunt. 2019.
\newblock Microsoft translator at wmt 2019: Towards large-scale document-level neural machine translation.
\newblock In \emph{Proceedings of the Fourth Conference on Machine Translation (Volume 2: Shared Task Papers, Day 1)}, pages 225--233.

\bibitem[{Kalyan et~al.(2021)Kalyan, Rajasekharan, and Sangeetha}]{xlm-r}
Katikapalli~Subramanyam Kalyan, Ajit Rajasekharan, and Sivanesan Sangeetha. 2021.
\newblock Ammus: A survey of transformer-based pretrained models in natural language processing.
\newblock \emph{arXiv preprint arXiv:2108.05542}.

\bibitem[{Kazemi et~al.(2021)Kazemi, Garimella, Gaffney, and Hale}]{kazemi2021claim}
Ashkan Kazemi, Kiran Garimella, Devin Gaffney, and Scott Hale. 2021.
\newblock Claim matching beyond english to scale global fact-checking.
\newblock In \emph{Proceedings of the 59th Annual Meeting of the Association for Computational Linguistics and the 11th International Joint Conference on Natural Language Processing (Volume 1: Long Papers)}, pages 4504--4517.

\bibitem[{King et~al.(2017)King, Butcher, and Zalewski}]{king_2017_438045}
Thomas King, Simon Butcher, and Lukasz Zalewski. 2017.
\newblock \href {https://doi.org/10.5281/zenodo.438045} {\emph{{Apocrita - High Performance Computing Cluster for Queen Mary University of London}}}.

\bibitem[{Lewoniewski et~al.(2024)Lewoniewski, Stolarski, Str{\'o}{\.z}yna, Lewa{\'n}ska, Wojewoda, Ksi{\k{e}}{\.z}niak, and Sawi{\'n}ski}]{lewoniewski2024openfact}
W{\l}odzimierz Lewoniewski, Piotr Stolarski, Milena Str{\'o}{\.z}yna, Elzbieta Lewa{\'n}ska, Aleksandra Wojewoda, Ewelina Ksi{\k{e}}{\.z}niak, and Marcin Sawi{\'n}ski. 2024.
\newblock Openfact at checkthat! 2024: Combining multiple attack methods for effective adversarial text generation.

\bibitem[{Li et~al.(2024)Li, Panchendrarajan, and Zubiaga}]{li2024factfinders}
Yufeng Li, Rrubaa Panchendrarajan, and Arkaitz Zubiaga. 2024.
\newblock Factfinders at checkthat! 2024: Refining check-worthy statement detection with llms through data pruning.

\bibitem[{Nakov et~al.(2022)Nakov, Barr{\'o}n-Cede{\~n}o, Da~San~Martino, Alam, M{\'\i}guez, Caselli, Kutlu, Zaghouani, Li, Shaar et~al.}]{nakov2022overview}
Preslav Nakov, Alberto Barr{\'o}n-Cede{\~n}o, Giovanni Da~San~Martino, Firoj Alam, Rub{\'e}n M{\'\i}guez, Tommaso Caselli, Mucahid Kutlu, Wajdi Zaghouani, Chengkai Li, Shaden Shaar, et~al. 2022.
\newblock Overview of the clef-2022 checkthat! lab task 1 on identifying relevant claims in tweets.
\newblock In \emph{2022 Conference and Labs of the Evaluation Forum, CLEF 2022}, pages 368--392. CEUR Workshop Proceedings (CEUR-WS. org).

\bibitem[{Panchendrarajan and Zubiaga(2024)}]{panchendrarajan2024claim}
Rrubaa Panchendrarajan and Arkaitz Zubiaga. 2024.
\newblock Claim detection for automated fact-checking: A survey on monolingual, multilingual and cross-lingual research.
\newblock \emph{Natural Language Processing Journal}, 7:100066.

\bibitem[{Panda and Levitan(2021)}]{panda2021detecting}
Subhadarshi Panda and Sarah~Ita Levitan. 2021.
\newblock Detecting multilingual covid-19 misinformation on social media via contextualized embeddings.
\newblock In \emph{Proceedings of the Fourth Workshop on NLP for Internet Freedom: Censorship, Disinformation, and Propaganda}, pages 125--129.

\bibitem[{Prabhakar et~al.(2020)Prabhakar, Mohtaj, and M{\"o}ller}]{prabhakar2020claim}
Acharya~Ashish Prabhakar, Salar Mohtaj, and Sebastian M{\"o}ller. 2020.
\newblock Claim extraction from text using transfer learning.
\newblock In \emph{Proceedings of the 17th International Conference on Natural Language Processing (ICON)}, pages 297--302.

\bibitem[{Rousseeuw(1987)}]{ROUSSEEUW198753}
Peter~J. Rousseeuw. 1987.
\newblock \href {https://doi.org/10.1016/0377-0427(87)90125-7} {Silhouettes: A graphical aid to the interpretation and validation of cluster analysis}.
\newblock \emph{Journal of Computational and Applied Mathematics}, 20:53--65.

\bibitem[{Savchev(2022)}]{savchev2022ai}
A~Savchev. 2022.
\newblock Ai rational at checkthat! 2022: using transformer models for tweet classification.
\newblock \emph{Working Notes of CLEF}.

\bibitem[{Sawi{\'n}ski et~al.(2023)Sawi{\'n}ski, W{\k{e}}cel, Ksi{\k{e}}{\.z}niak, Str{\'o}{\.z}yna, Lewoniewski, Stolarski, and Abramowicz}]{sawinski2023openfact}
Marcin Sawi{\'n}ski, Krzysztof W{\k{e}}cel, Ewelina Ksi{\k{e}}{\.z}niak, Milena Str{\'o}{\.z}yna, W{\l}odzimierz Lewoniewski, Piotr Stolarski, and Witold Abramowicz. 2023.
\newblock Openfact at checkthat! 2023: Head-to-head gpt vs. bert-a comparative study of transformers language models for the detection of check-worthy.

\bibitem[{Sinaga and Yang(2020)}]{sinaga2020unsupervised}
Kristina~P Sinaga and Miin-Shen Yang. 2020.
\newblock Unsupervised k-means clustering algorithm.
\newblock \emph{IEEE access}, 8:80716--80727.

\bibitem[{Suri and Dudeja(2022)}]{suri2022asatya}
PK~Manan Suri and S~Dudeja. 2022.
\newblock Asatya at checkthat! 2022: multimodal bert for identifying claims in tweets.
\newblock \emph{Working Notes of CLEF}.

\bibitem[{Team et~al.(2022)Team, Costa-jussà, Cross, Çelebi, Elbayad, Heafield, Heffernan, Kalbassi, Lam, Licht, Maillard, Sun, Wang, Wenzek, Youngblood, Akula, Barrault, Gonzalez, Hansanti, Hoffman, Jarrett, Sadagopan, Rowe, Spruit, Tran, Andrews, Ayan, Bhosale, Edunov, Fan, Gao, Goswami, Guzmán, Koehn, Mourachko, Ropers, Saleem, Schwenk, and Wang}]{nllb2022}
NLLB Team, Marta~R. Costa-jussà, James Cross, Onur Çelebi, Maha Elbayad, Kenneth Heafield, Kevin Heffernan, Elahe Kalbassi, Janice Lam, Daniel Licht, Jean Maillard, Anna Sun, Skyler Wang, Guillaume Wenzek, Al~Youngblood, Bapi Akula, Loic Barrault, Gabriel~Mejia Gonzalez, Prangthip Hansanti, John Hoffman, Semarley Jarrett, Kaushik~Ram Sadagopan, Dirk Rowe, Shannon Spruit, Chau Tran, Pierre Andrews, Necip~Fazil Ayan, Shruti Bhosale, Sergey Edunov, Angela Fan, Cynthia Gao, Vedanuj Goswami, Francisco Guzmán, Philipp Koehn, Alexandre Mourachko, Christophe Ropers, Safiyyah Saleem, Holger Schwenk, and Jeff Wang. 2022.
\newblock No language left behind: Scaling human-centered machine translation.

\bibitem[{Tedeschi and Navigli(2022)}]{tedeschi-navigli-2022-multinerd}
Simone Tedeschi and Roberto Navigli. 2022.
\newblock \href {https://doi.org/10.18653/v1/2022.findings-naacl.60} {{M}ulti{NERD}: A multilingual, multi-genre and fine-grained dataset for named entity recognition (and disambiguation)}.
\newblock In \emph{Findings of the Association for Computational Linguistics: NAACL 2022}, pages 801--812, Seattle, United States. Association for Computational Linguistics.

\bibitem[{Uyangodage et~al.(2021)Uyangodage, Ranasinghe, and Hettiarachchi}]{uyangodage2021can}
Lasitha Uyangodage, Tharindu Ranasinghe, and Hansi Hettiarachchi. 2021.
\newblock Can multilingual transformers fight the covid-19 infodemic?
\newblock In \emph{Proceedings of the International Conference on Recent Advances in Natural Language Processing (RANLP 2021)}, pages 1432--1437.

\bibitem[{Vaswani et~al.(2017)Vaswani, Shazeer, Parmar, Uszkoreit, Jones, Gomez, Kaiser, and Polosukhin}]{vaswani2017attention}
Ashish Vaswani, Noam Shazeer, Niki Parmar, Jakob Uszkoreit, Llion Jones, Aidan~N Gomez, {\L}ukasz Kaiser, and Illia Polosukhin. 2017.
\newblock Attention is all you need.
\newblock \emph{Advances in neural information processing systems}, 30.

\bibitem[{Zeng et~al.(2021)Zeng, Abumansour, and Zubiaga}]{zeng2021automated}
Xia Zeng, Amani~S Abumansour, and Arkaitz Zubiaga. 2021.
\newblock Automated fact-checking: A survey.
\newblock \emph{Language and Linguistics Compass}, 15(10):e12438.

\bibitem[{Zengin et~al.(2021)Zengin, Kartal, and Kutlu}]{zengin2021tobb}
Muhammed~Said Zengin, Yavuz~Selim Kartal, and M{\"u}cahid Kutlu. 2021.
\newblock Tobb etu at checkthat! 2021: Data engineering for detecting check-worthy claims.
\newblock In \emph{CLEF (Working Notes)}, pages 670--680.

\end{thebibliography}

\appendix

\section{Statistics}\label{appendix:statistics}

\begin{figure*}[!t]
\centering    
\includegraphics[width=0.8\textwidth]{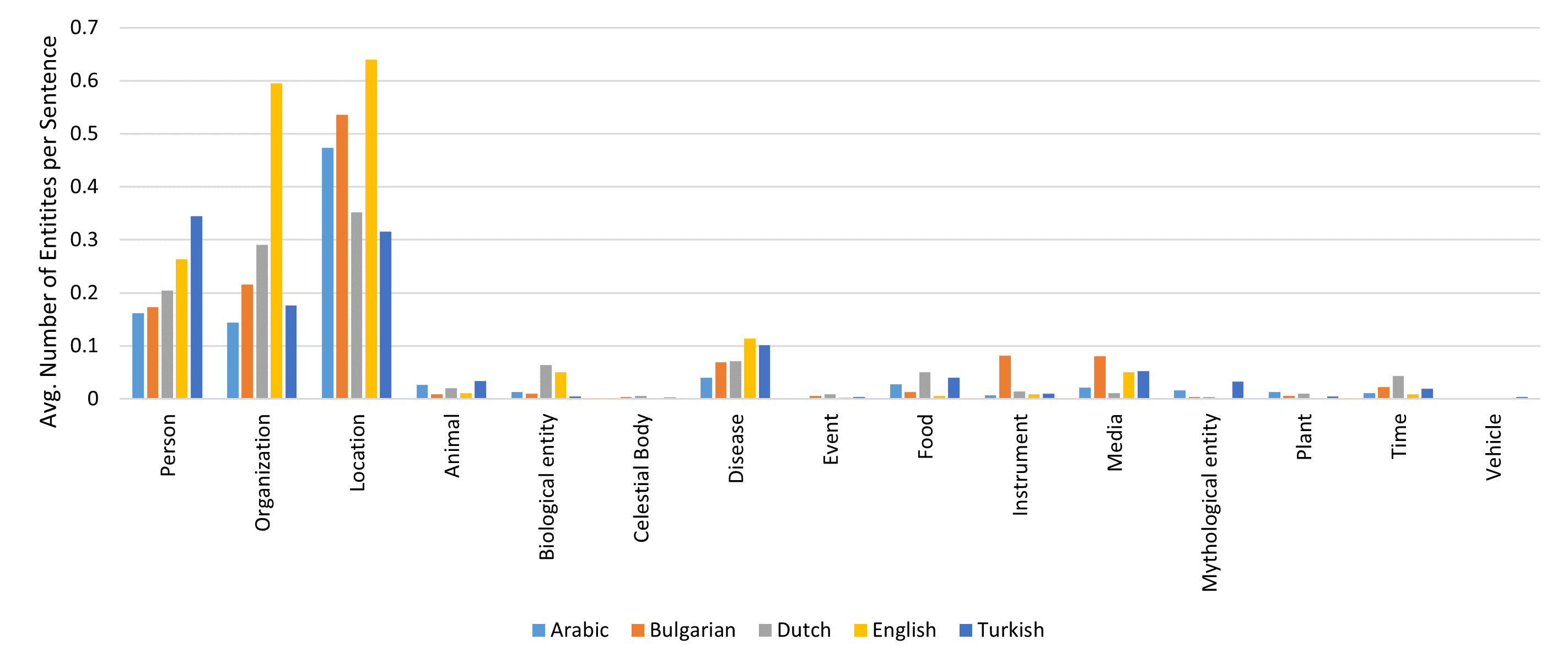}
\caption{Average Number of Named Entities Identified in the Train Partition of CheckThat! 2022 Data}
\label{fig:EX-Claim-F_stat}
\end{figure*}

\subsection{Statistics of Named Entities Recognized}\label{appendix:NER_stats}
MultiNERD identifies the following 15 types of fine-grained Named Entities.
\begin{compactitem}
    \item Person (PER)
    \item Organization (ORG)
    \item Location (LOC)
    \item Animal (ANIM)
    \item Biological entity (BIO)
    \item Celestial Body (CEL)
    \item Disease (DIS)
    \item Event (EVE)
    \item Food (FOOD)
    \item Instrument (INST)
    \item Media (MEDIA)
    \item Plant (PLANT)
    \item Mythological entity (MYTH)
    \item Time (TIME)
    \item Vehicle (VEHI)
\end{compactitem}

Figures \ref{fig:EX-Claim-F_stat} show the average number of named entities recognized in the Train partition of CheckThat! 2022 data. Notably, \textit{location} entities appear more frequently across all languages. This could be due to the COVID-19 discussion in the training data referring to the status of the disease from different countries and locations. Additionally, the occurrences of \textit{disease}, \textit{media}, \textit{biological entities}, and \textit{food} are relatively higher compared to other rare entity types, likely due to the COVID-19-related discussions present in the dataset. 

\subsection{Statistics of Entities Linked}\label{appendix:entity_linking_stats}

\begin{figure}[t]
    \centering 
    \includegraphics[width=0.47\textwidth]{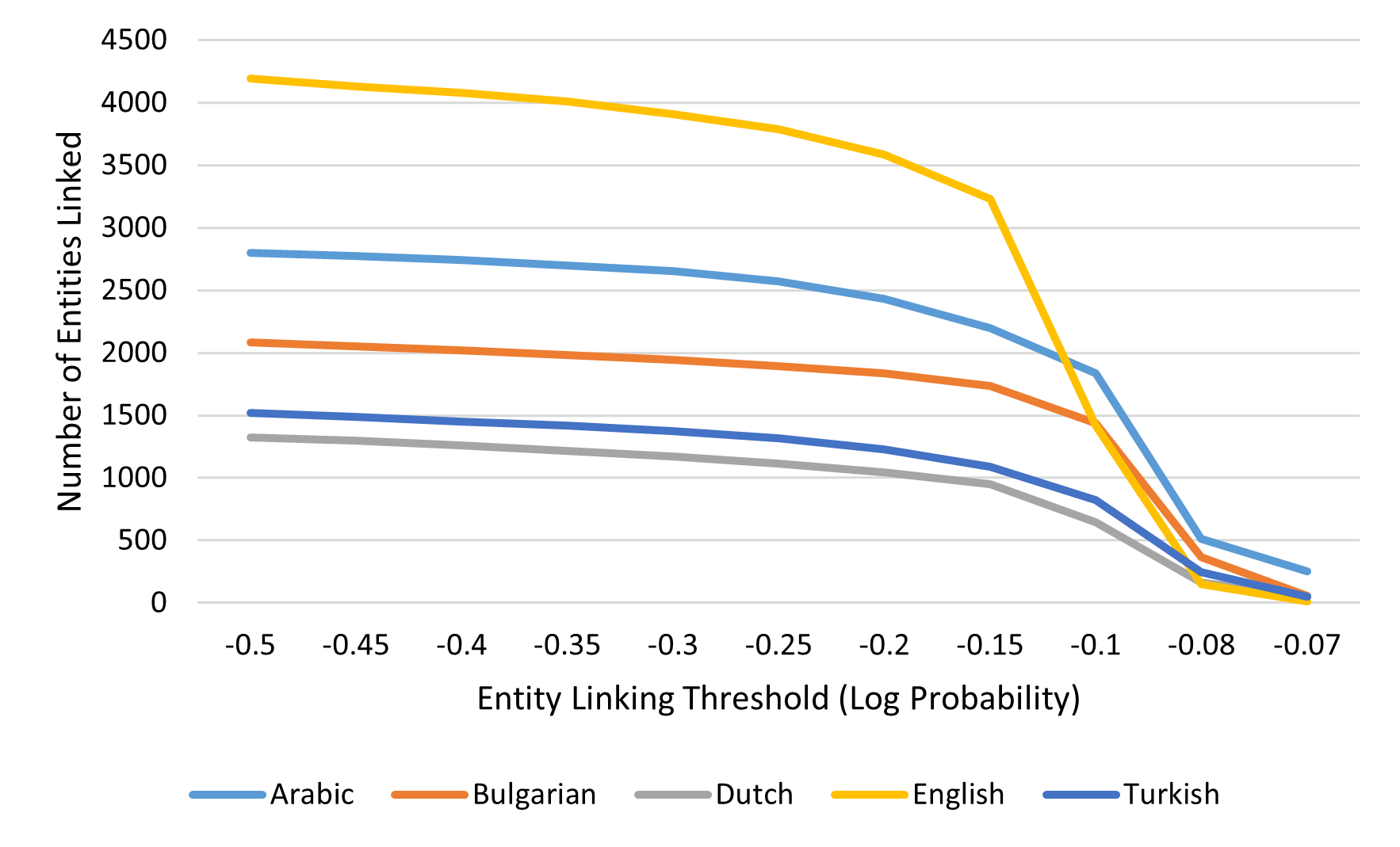}
    \caption{Number of Entities Linked with the Variation in EL Threshold in Train Partition of CheckThat! Data}
    \label{fig:WikiNER_Link_stat}
\end{figure} 

\begin{figure}[t]
    \centering
    \includegraphics[width=0.39\textwidth]{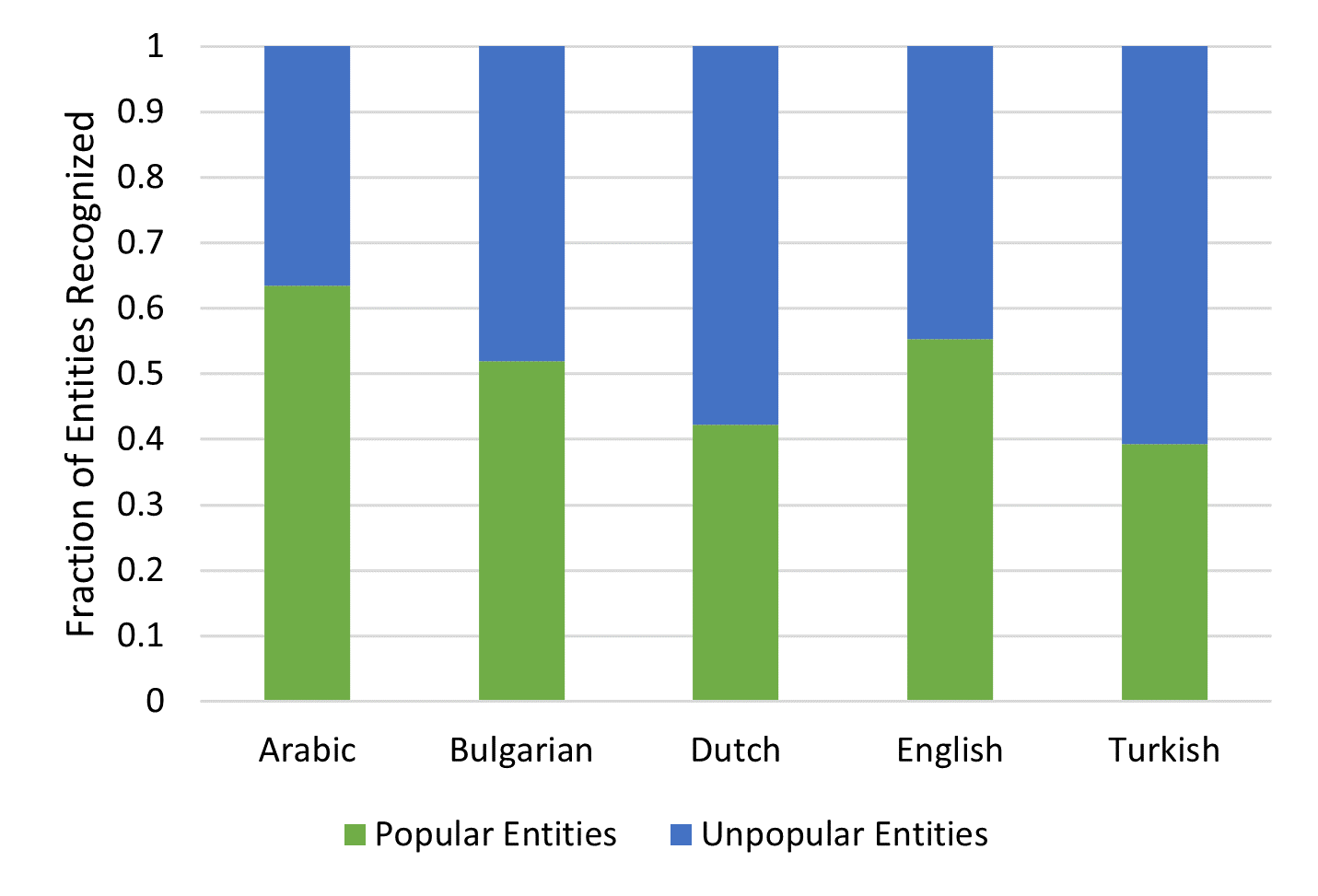}
    \caption{Popular vs Unpopular Fine-grained Entities}
    \caption{Fraction of Popular and Unpopular Entities in Train Partition of CheckThat!  Data}
    \label{fig:WikiNER_stat}
\end{figure}

Figure \ref{fig:WikiNER_Link_stat} shows the number of named entities linked to a Wikipedia page with the increase in threshold. It can be observed that the reduction in the number of entities linked to a Wikipedia page with the increase in log probability threshold is not very significant until the threshold is increased to -0.15 (probability score of 0.861). This indicates that most of the NER identified are linked to a Wikipedia page with a very high probability. 

Figure \ref{fig:WikiNER_stat} shows the fraction of popular and unpopular entities determined with the log probability threshold of -0.15 (probability of 0.861). Notably, Arabic language data contains more fraction of entities identified as popular followed by English, and Dutch and Turkish have the lowest fraction among all. Further, the chances of fine-grained entities being popular range from 0.39 to 0.63, highlighting that nearly more than 40\% of the entities recognized are linked as popular ones with a very high probability.

\subsection{Statistics of the Datasets Used}\label{sec:dataset_stats}

\subsubsection{CheckThat! Dataset}\label{appendix_checkthat_stats}

\begin{table*}[t]
\centering
\small
\begin{tabular}{llllllll}\hline
                          &       & Arabic & Bulgarian & Dutch & English & Turkish & Total                  \\ \hline 
\multirow{3}{*}{Train}    & Verifiable   & 2,513   & 1,871      & 929   & 2,122    & 1,589     & \multirow{3}{*}{14,032} \\
                          & Unverifiable    & 1,118   & 839       & 1,021  & 1,202    & 828      &                         \\
                          & Total & 3,631   & 2,710      & 1,950  & 3,324    & 2,417     &                         \\ \hline
\multirow{3}{*}{Dev}      & Verifiable   & 235    & 177       & 72    & 195     & 150      & \multirow{3}{*}{1,300}   \\
                          & Unverifiable    & 104    & 74        & 109   & 112     & 72       &                         \\
                          & Total & 339    & 251       & 181   & 307     & 222      &                         \\ \hline
\multirow{3}{*}{Dev-Test} & Verifiable   & 691    & 519       & 252   & 574     & 438      & \multirow{3}{*}{3,837}   \\
                          & Unverifiable    & 305    & 217       & 282   & 337     & 222      &                         \\
                          & Total & 996    & 736       & 534   & 911     & 660      &                         \\ \hline
\multirow{3}{*}{Test}     & Verifiable   & 682    & 199       & 608   & 149     & 303      & \multirow{3}{*}{3,698}   \\
                          & Unverifiable    & 566    & 130       & 750   & 102     & 209      &                         \\
                          & Total & 1,248   & 329       & 1,358  & 251     & 512      & \\ \hline                        
\end{tabular}
\caption{Statistics of the CheckThat! 2022 Verifiable Claim Detection Dataset}\label{tab:checkthat_2022_stats}
\end{table*}

Table \ref{tab:checkthat_2022_stats} presents the statistics of the CheckThat! 2022 dataset. Compared to other languages, the number of data instances is low in Dutch across all partitions except Test. Further, the ratio between verifiable and unverifiable claims is equal for Dutch languages, whereas other languages contain more verifiable statements than unverifiable (roughly 2:1 ratio).

\subsubsection{Kazemi Dataset}\label{appendix_kazemi_stats}

\begin{table*}[t]
\small
\centering
\begin{tabular}{lllllll} \hline
      & Bengali & English & Hindi & Malayalam & Tamil & Total                  \\ \hline
Verifiable   & 329     & 535     & 407   & 707       & 197   & \multirow{3}{*}{4,017} \\
Unverifiable    & 593     & 388     & 465   & 263       & 133   &                        \\
Total & 922     & 923     & 872   & 970       & 330   &                       \\ \hline
\end{tabular}
\caption{Statistics of the Kazemi 2021 Verifiable Claim Detection Dataset}\label{tab:kazemi_2021_stats}
\end{table*}

Table \ref{tab:kazemi_2021_stats} presents the statistics of the Kazemi dataset. The dataset contains WhatsApp tiplines and social media posts about topics such as politics and COVID-19. 

\section{Results}

\begin{table*}[t]
\centering
\small
\begin{tabular}{l|cccccc|ccc}\hline
& \multicolumn{6}{c|}{Baseline} & \multicolumn{3}{c}{Entity Type Categories}    \\ \hline
Language   & \rotatebox{90}{mBERT}         & \rotatebox{90}{XLM-R}   & \rotatebox{90}{mT5}     & \rotatebox{90}{Llama2}         & \rotatebox{90}{Mistral}   & \rotatebox{90}{Phi3} & \rotatebox{90}{X-Claim}  & \rotatebox{90}{EXN-Claim}  & \rotatebox{90}{EXP-Claim} \\ \hline
Arabic     & 0.692 & 0.654 & 0.677 & \textbf{0.752} & 0.736 & 0.718 & 0.731 & 0.733 & 0.735 \\
Bengali    & 0.725 & 0.655 & 0.556 & 0.697 & 0.642 & 0.552 & 0.713 & 0.748 & \textbf{0.764} \\
Bulgarian  & 0.783 & 0.633 & 0.729 & \textbf{0.835} & 0.815 & 0.812 & 0.782 & 0.79  & 0.796 \\
Czech      & 0.707 & 0.647 & 0.72 & \textbf{0.808} & 0.786 & 0.755 & 0.741 & 0.774 & 0.769 \\
Dutch      & 0.68  & 0.698 & 0.634 & 0.72 & 0.657 & 0.663 & 0.716 & 0.717 & \textbf{0.718} \\
English (Kaz)  & 0.738 & \textbf{0.769} & 0.748 & 0.764 & 0.75 & 0.719 & 0.755 & 0.757 & 0.755 \\
English (CT)   & 0.724 & 0.658 & 0.688 & 0.769 & \textbf{0.777} & 0.755 & 0.75  & \textbf{0.762} & 0.761 \\
French     & 0.638 & 0.698 & 0.658 & 0.736 & \textbf{0.739} & 0.738 & 0.703 & 0.722 & 0.725 \\
German     & 0.751 & 0.714 & 0.722 & 0.808 & \textbf{0.822} & 0.816 & 0.789 & 0.778 & 0.775 \\
Greek      & 0.677 & 0.64  & 0.71 & 0.697 & 0.687 & 0.679 & 0.735 & 0.747 & \textbf{0.759} \\
Gujarati   & 0.661 & 0.669 & 0.649 & 0.457 & 0.582 & 0.455 & \textbf{0.77} & \textbf{0.77}  & 0.751 \\
Hindi      & 0.631 & 0.695 & 0.613 & 0.792 & \textbf{0.809} & 0.6 & 0.766 & 0.767 & 0.769 \\
Hungarian  & 0.667 & 0.651 & 0.665 & \textbf{0.757} & 0.727 & 0.637 & 0.68  & 0.683 & 0.682 \\
Italian    & 0.721 & 0.647 & 0.704 & \textbf{0.764} & 0.745 & 0.736 & 0.709 & 0.731 & 0.723 \\
Japanese   & 0.716 & 0.658 & 0.693 & 0.759 & \textbf{0.771} & 0.74 & 0.715 & 0.732 & 0.746 \\
Kannada    & 0.535 & 0.567 & 0.579 & 0.371 & 0.459 & 0.342 & 0.628 & 0.674 & \textbf{0.676} \\
Korean     & 0.691 & 0.635 & 0.623 & 0.69 & 0.72 & 0.653 & 0.712 & \textbf{0.718} & 0.709 \\
Marathi    & 0.588 & 0.64  & 0.603 & 0.643 & 0.532 & 0.46 & \textbf{0.668} & 0.666 & 0.664 \\
Malayalam  & 0.79  & 0.564 & 0.685 & 0.374 & 0.151 & 0.141 & \textbf{0.845} & 0.84  & 0.835 \\
Persian    & 0.673 & 0.669 & 0.622 & 0.713 & 0.667 & 0.602 & 0.704 & 0.716 & \textbf{0.718} \\
Russian    & 0.709 & 0.667 & 0.641 & \textbf{0.766} & 0.764 & 0.734 & 0.715 & 0.724 & 0.724 \\
Spanish    & 0.738 & 0.693 & 0.683 & 0.779 & 0.735 & 0.77 & \textbf{0.787} & 0.783 & 0.784 \\
Swedish    & 0.707 & 0.688 & 0.654 & 0.772 & 0.779 & 0.76 & 0.777 & 0.778 &  \textbf{0.787} \\
Tamil      & 0.623 & 0.744 & 0.789 & 0.439 & 0.398 & 0.264 & 0.864 & \textbf{0.866} & 0.861 \\
Thai       & 0.628 & 0.648 & 0.644 & 0.716 & 0.698 & 0.673 & \textbf{0.726} & 0.716 & 0.715 \\
Turkish    & 0.699 & 0.648 & 0.664 & 0.775 & \textbf{0.82} & 0.745 & 0.748 & 0.759 & 0.758 \\
Ukrainian  & 0.688 & 0.597 & 0.604 & \textbf{0.764} & 0.724 & 0.719 & 0.644 & 0.688 & 0.707 \\
Vietnamese & 0.727 & 0.672 & 0.676 & \textbf{0.735} & 0.694 & 0.689 & 0.715 & 0.717 & 0.729 \\ \hline 
\end{tabular}
\caption{F1-Score of the Claim Detection Models at Language-level}\label{tab:f1_score_languages}
\end{table*}

\subsection{Language-level Performance}\label{sec:laguage_level_performance}
Table \ref{tab:f1_score_languages} lists the language-level F1-score of the claim detection models across 27 languages present in the test datasets. 

\subsection{Machine Translation Performance}\label{sec:machine_translation_performance}
Since the synthetic dataset was generated using machine translation, we assess the quality of the translation model—Microsoft Azure AI Translator \cite{junczys2019microsoft}. For this evaluation, we utilize FLORES-200 \cite{nllb2022}, a multilingual benchmark dataset that supports English-to-200+ language translation. To approximate the conditions of our synthetic dataset, we sampled sentence pairs from the five source languages represented in the CheckThat! training set to 18 target languages present in the synthetic dataset. For each target language, we selected 250 translation pairs, resulting in a total of 4,500 sentence pairs. This sampling strategy is designed to mimic the structure of our synthetic dataset, where each target language includes 250 translated claims.

\begin{table}[]
\small
\centering
\begin{tabular}{ll}\hline
Metric     & Score \\ \hline
BLEU      & 26.23 \\
CHRF      & 54.35  \\
CHRF++ & 51.63  \\ \hline                                  \end{tabular}
\caption{Machine Translation Performance of Microsoft Azure AI translator}\label{tab:ml_performance}
\end{table}

Table \ref{tab:ml_performance} presents the performance of the machine translation model on the sampled translation pairs, evaluated using multiple translation quality metrics. The model achieves competitive scores on the character-based metrics CHRF and CHRF++, with the CHRF++ score notably surpassing the baseline results reported in the original FLORES-200 dataset paper \cite{nllb2022}.

\section{Analysis}
\subsection{Entity-type Embedding}

\begin{figure}[t]
    \centering
    \includegraphics[width=0.49\textwidth]{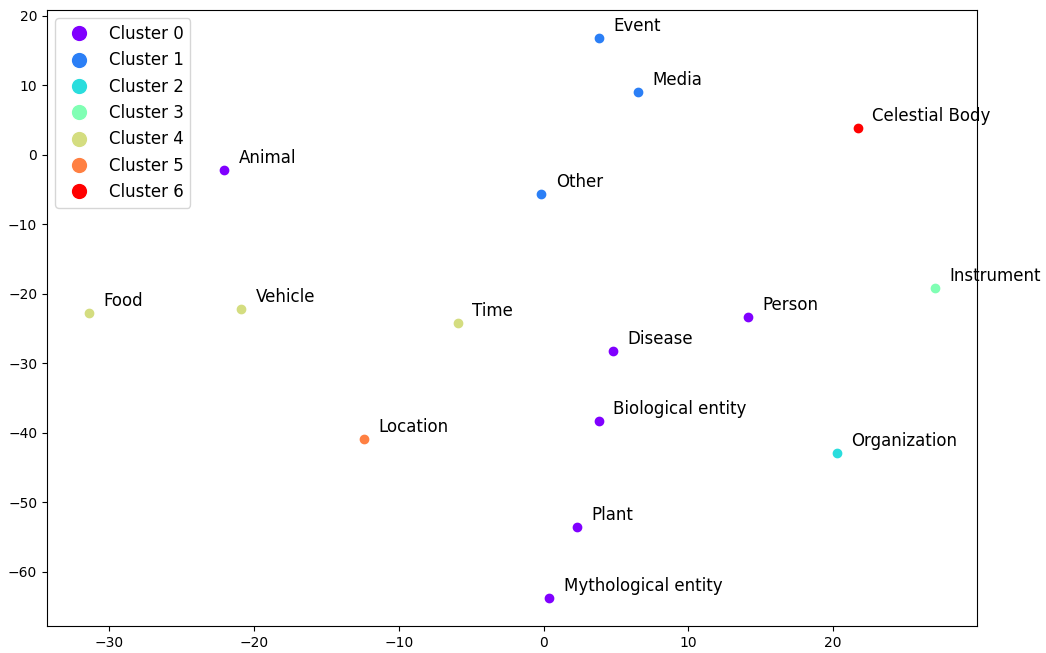}
    \caption{Two-dimensional embedding visualization of EXN-Claim Model}
    \label{fig:embedding_visualization}
\end{figure}

Unlike the cross-lingual word embedding which is directly obtained from the pre-trained model, \textit{EX-Claim} learns a dense vector representation for each entity type. Figure \ref{fig:embedding_visualization} shows the 2D visualization of the entity type embeddings learned by the \textit{EXN-Claim}  model. The 2D vector representation of the embeddings is obtained using the t-SNE\footnote{\url{https://scikit-learn.org/stable/modules/generated/sklearn.manifold.TSNE.html}} dimension reduction technique. Further, the embedding representations were clustered using the KMeans clustering technique \citep{sinaga2020unsupervised}, and the number of optimal clusters was chosen using the Silhouette score \citep{ROUSSEEUW198753}.

 The \textit{EXN-Claim} model learns various groups of entity types. Notably, this model clusters \textit{Animal, Disease, Person, Biological Entity, Disease, Plan} and \textit{Mythological Entity} into a single group, likely influenced by the COVID-19 discussions present in the training data. This observation highlights the model's tendency to learn context-specific entity-type embeddings during training.

\subsection{Attention Weights for Sample Sentences}
\label{sec:appendix_attention_weights}

\begin{figure*}[t]
    \centering
    \begin{subfigure}[b]{0.49\textwidth}
    \includegraphics[width=\textwidth]{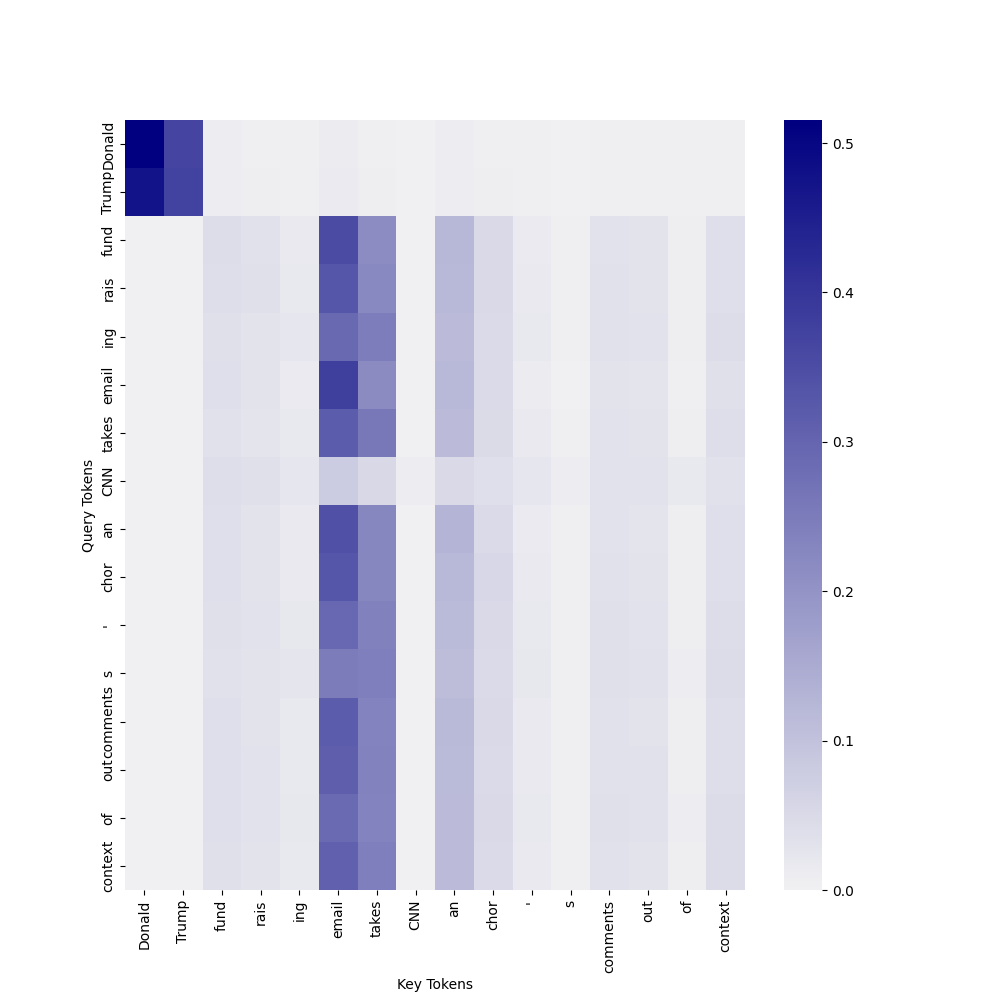}
    \caption{S4 : \textit{EXN-Claim}}
    \end{subfigure}  
    \begin{subfigure}[b]{0.4\textwidth}
    \includegraphics[width=\textwidth]{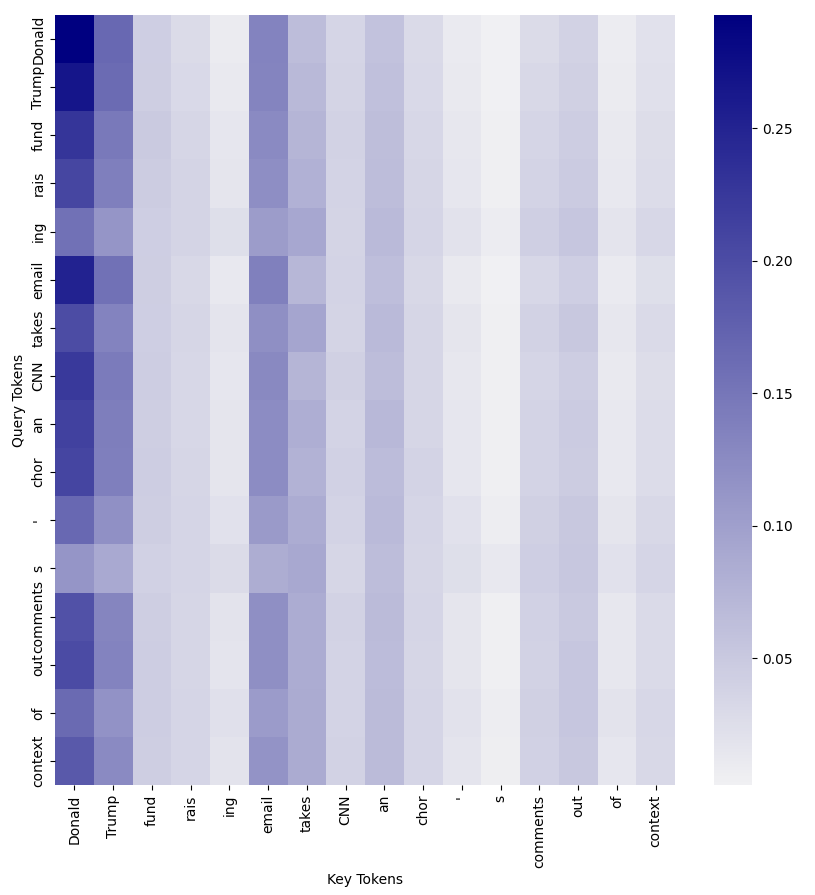}
    \caption{S4 : \textit{X-Claim}}  
    \end{subfigure}
    \begin{subfigure}[b]{0.44\textwidth}
    \includegraphics[width=\textwidth]{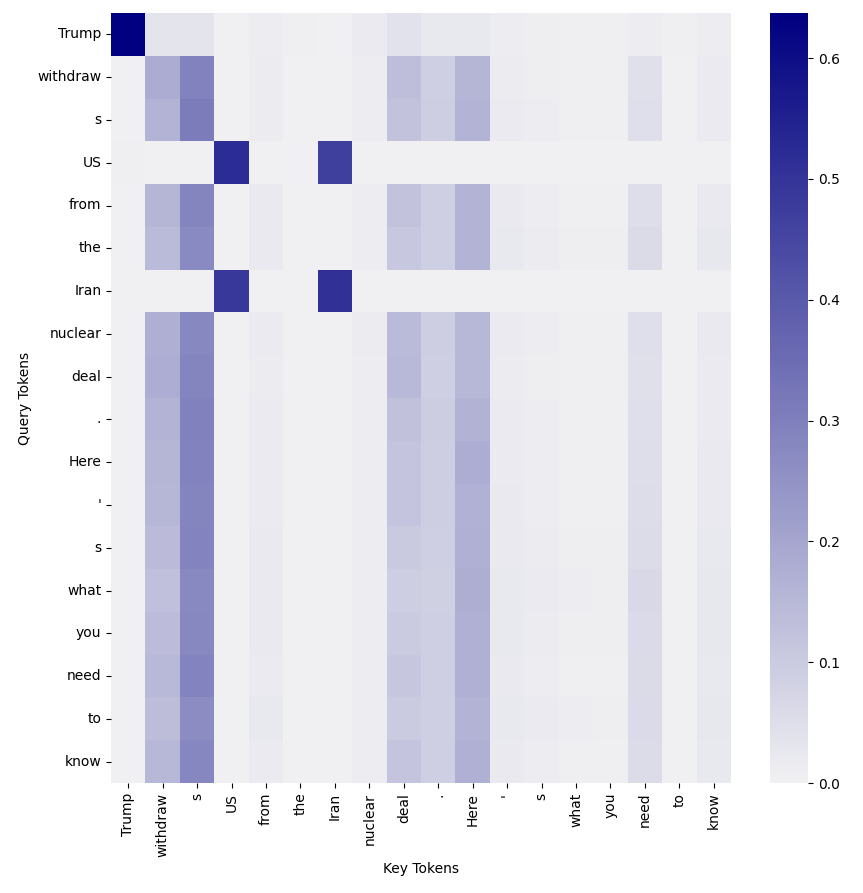}
    \caption{S11 : \textit{EXN-Claim}}  
    \end{subfigure}  
    \begin{subfigure}[b]{0.44\textwidth}
    \includegraphics[width=\textwidth]{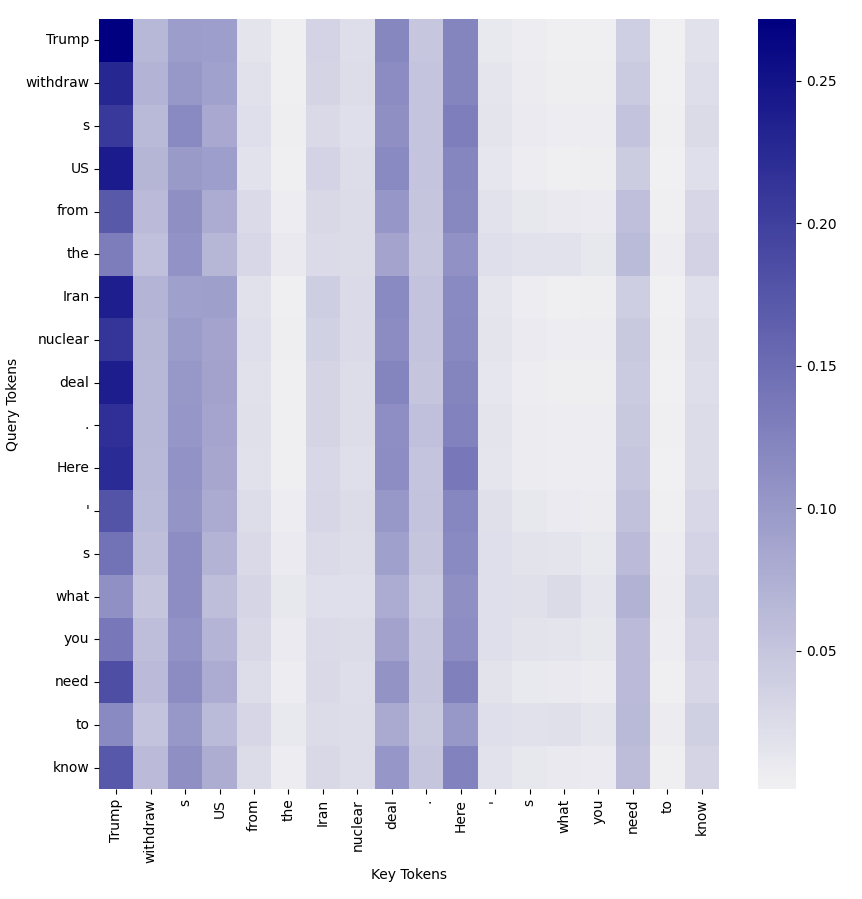}
    \caption{S11 : \textit{X-Claim}}   
    \end{subfigure}
    \begin{subfigure}[b]{0.41\textwidth}
    \includegraphics[width=\textwidth]{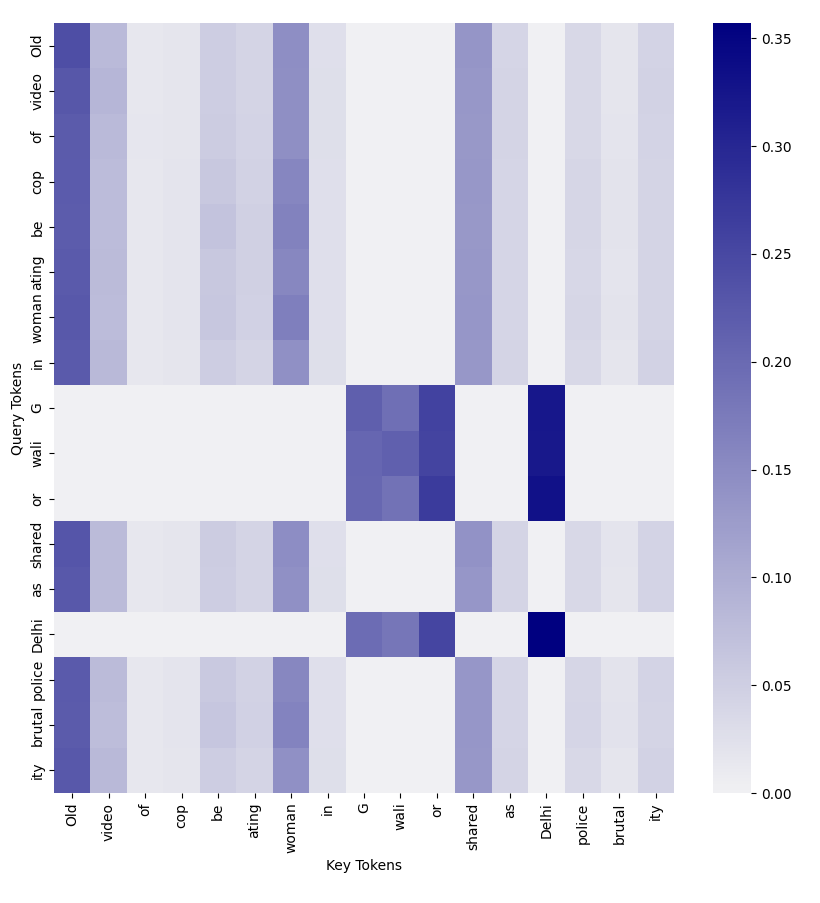}
    \caption{S12 : \textit{EXP-Claim}}  
    \end{subfigure}  
    \begin{subfigure}[b]{0.41\textwidth}
    \includegraphics[width=\textwidth]{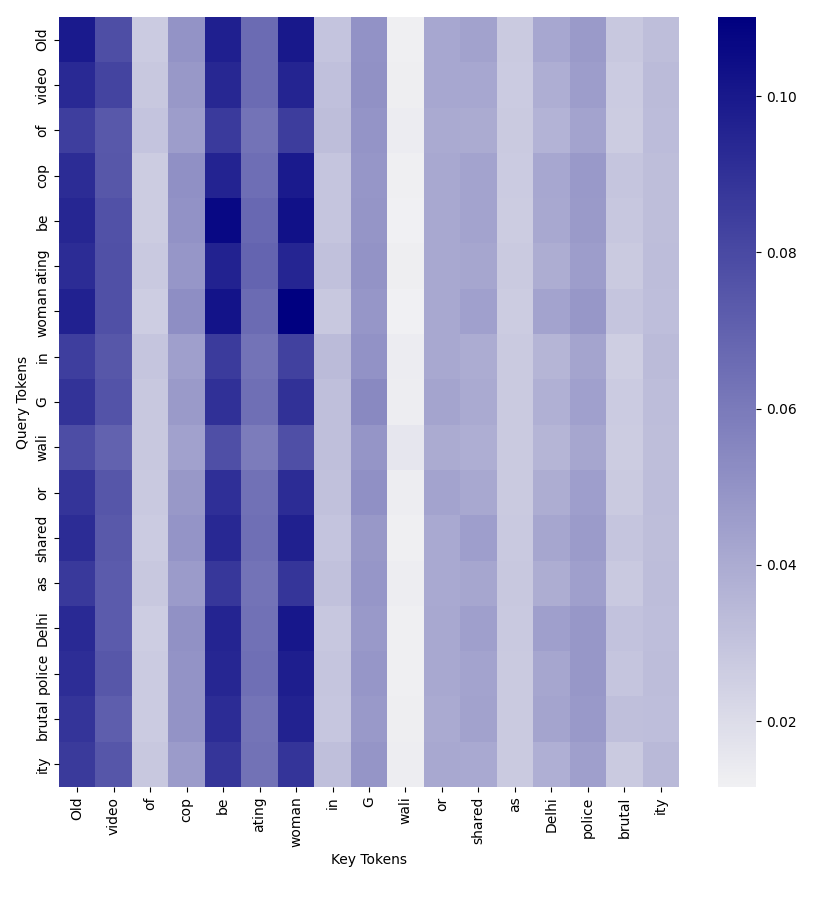}
    \caption{S12 : \textit{X-Claim}}
    \end{subfigure}
    \caption{Heatmap of Attention Weights Generated by Claim Detection Models for sentences S10, S11, and S12}
    \label{fig:attention_heatmaps}
\end{figure*}

Figure \ref{fig:attention_heatmaps} shows the heatmaps of the attention weights computed by different claim detection models for the following three sentences.

\begin{compactitem}
    \item \textit{S10} - \textit{Old video of cop beating woman in \textbf{Gawalior} shared as \textbf{Delhi} police brutality}
    \item \textit{S11} - \textit{\textbf{Donald Trump} fund raising email takes \textbf{CNN} anchor's comments out of context}
    \item \textit{S12} - \textit{\textbf{Trump} withdraws \textbf{US} from the \textbf{Iran} nuclear deal}
\end{compactitem}

It can be observed that \textit{EX-Claim} attends to important words in the sentences, whereas \textit{X-Claim} generates attention weights distributed across most of the words. Further, an interesting pattern of named entities attending to themselves or other named entities more compared to other words can be observed across all three sentences.

\subsection{Ablation Study}
We conduct an ablation study to identify the impact of the key components of \textit{EX-Claim} with the following variations.
\begin{compactitem}
    \item No projection layer (NPL) - The projection layer is omitted, and attention is applied directly to the embedding representations.
    \item No attention layer (NAL) - The attention layer is removed, hence the projected vector representations are aggregated into a sentence embedding by computing their mean. 
    \item No embedding layer (NEL) - The embedding layer is removed, and the entity information is represented as one-hot vectors.
\end{compactitem}

\begin{table}[H]
\small
\centering
\begin{tabular}{ll}\hline
Model    & F1-Score \\ \hline
X-Claim      & \textbf{0.735} \\
X-Claim - NPL      & 0.688 \\
X-Claim - NAL  & 0.714 \\\hline                                  \end{tabular}
\caption{Ablation Results of \textit{X-Claim} Model}\label{tab:ablation_none}
\end{table}

The \textit{NLP} and \textit{NAL} variations are applied to \textit{X-Claim}, and Table \ref{tab:ablation_none} presents the overall F1-score of these models. The results demonstrate that the projection layer is crucial for learning task-specific input representations, as its removal leads to a 4.7\% drop in overall performance. Likewise, the attention layer enhances the model's performance, contributing to a 2.1\% improvement.

\begin{table}[H]
\small
\centering
\begin{tabular}{ll}\hline
Model    & F1-Score \\ \hline
EXN-Claim      & \textbf{0.745} \\
EXN-Claim - NEL      & 0.733 \\ \hline
EXP-Claim      & \textbf{0.746} \\
EXP-Claim- - NEL      & 0.733 \\ \hline \end{tabular}
\caption{Ablation Results of \textit{EX-Claim Models}}\label{tab:ablation_entity}
\end{table}

Table \ref{tab:ablation_entity} lists the performance of \textit{EX-Claim} models with \textit{NEL} variation. It can be seen, that the performance declines by approximately 1.1\% - 1.3\% when entity information is represented as one-hot vectors, suggesting that the models struggle to effectively utilize entity information when it is provided as a one-hot vector representation.

\subsection{Correlation between Confusion Matrix and Selected Features}\label{appendix:correlation_matrix}

\begin{figure}[H]
    \centering
    \includegraphics[width=0.38\textwidth]{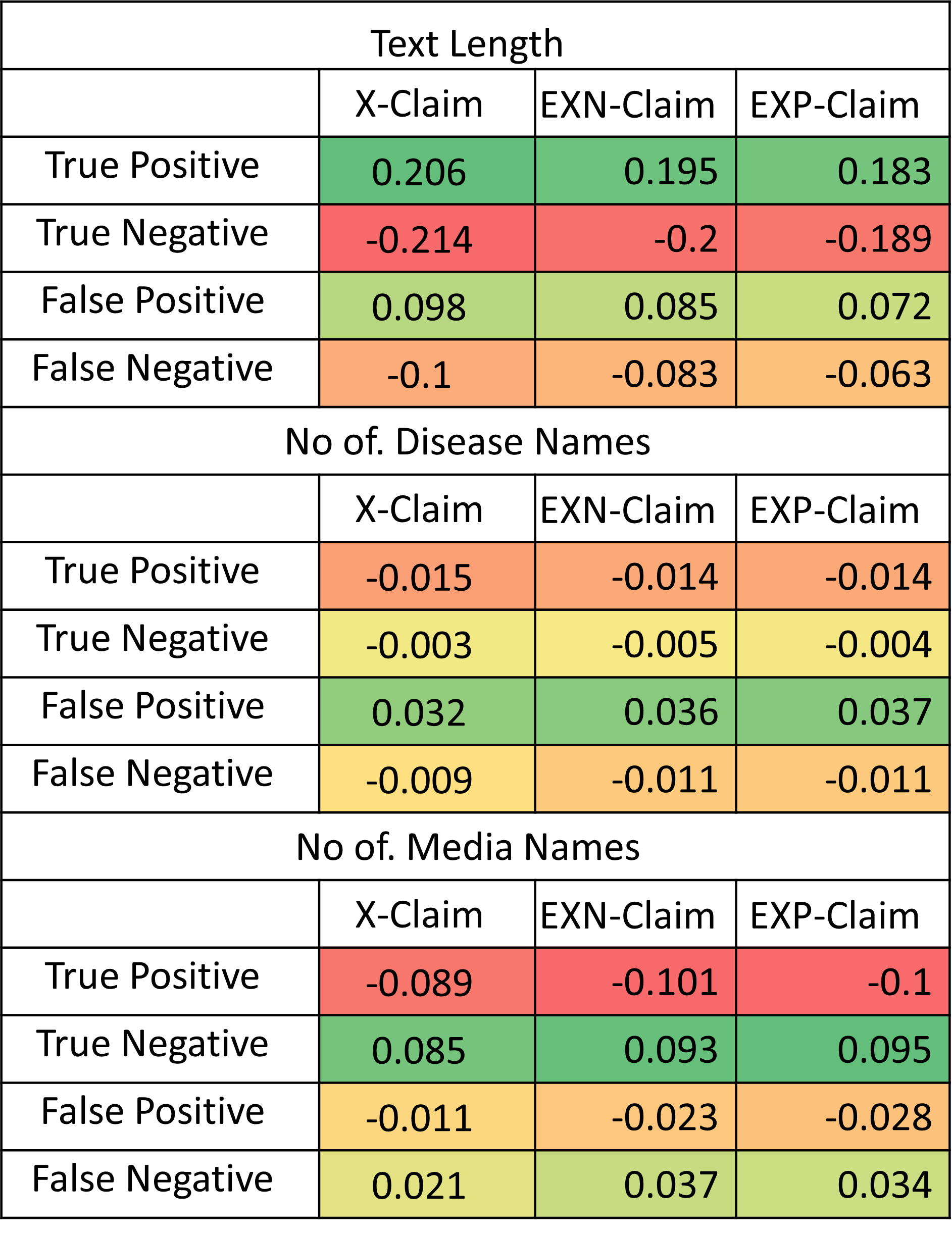}
    \caption{Correlation of Sample Features vs Confusion Matrix}
    \label{fig:correlation_matrix}
\end{figure}

Figure \ref{fig:correlation_matrix} presents the correlation scores between the true positive, true negative, false positive, and false negative values and selected features. It can be observed that the \textit{length} of the claim is positively correlated with true positive and false positive rates, showing the models tend to assume lengthy claims as verifiable.

Similarly, the number of disease names appearing in a claim is positively correlated with a false positive rate. This could be the COVID-19 discussion in the training data making the model to classify the claims with disease names as verifiable. Further, the number of media names is positively correlated with  both true negative and false negative rates as discussed in Section \ref{sec:error_analysis}. This suggests that references to media, such as images and videos, can either enhance or hinder verifiability, depending on the context and reach. 

\section{Implementation}\label{sec:implementation}

\begin{table*}[t]
\centering
\small
\begin{tabular}{@{}llccc@{}}
\hline
\textbf{} & \textbf{Model} & \textbf{\# Parameters} & \textbf{Multilingual?} & \textbf{Training time} \\
\hline
\multirow{3}{*}{\begin{tabular}[c]{@{}l@{}}\textbf{Transformer-based}\\ \textbf{Baselines}\end{tabular}} 
& mBERT               & 110M     & \checkmark\ Yes (104 langs)            & 20 sec \\
& XLM-R               & 270M     & \checkmark\ Yes (100 langs)            & 20 sec \\
& mT5 Large Encoder   & 600M     & \checkmark\ Yes (101 langs)            & 20 sec \\
\hline
\multirow{3}{*}{\textbf{LLMs}} 
& LLaMA 2             & 7B      & Limited                      & 3 hours \& 16 mins \\
& Mistral             & 7B     & Limited                      & 3 hours \& 25 mins \\
& Phi-3               & 3.8B   &  Limited                      & 2 hours \& 12 mins \\
\hline
\multirow{3}{*}{\textbf{Entity-aware Models}} 
& \textit{X-Claim}            & 197K     & \checkmark\ Yes (100 langs)            & 20 sec \\
& \textit{EXN-Claim}            & 201K     & \begin{tabular}[c]{@{}l@{}}\checkmark\ Yes (100 langs\\  + NER model langs) \end{tabular} & 20 sec \\
& \textit{EXP-Claim}           & 206K     & \begin{tabular}[c]{@{}l@{}}\checkmark\ Yes (100 langs\\  + NER model langs \\ \& EL model langs) \end{tabular} & 21 sec \\
\hline
\end{tabular}
\caption{Comparison of Model Parameters, Multilingual Support, and Training Time}\label{tab:efficiency_comparison}
\end{table*}

\subsection{Resources}
All the experiments were conducted using Queen Mary's Apocrita HPC facility, supported by QMUL Research-IT \cite{king_2017_438045}. Specifically, 1 GPU (Volta V100 or Ampere A100) with 8 CPU cores, each composed of 11 GB memory was used to train and test all the models.

\subsection{Models}
We fine-tuned the mBERT\footnote{https://huggingface.co/google-bert/bert-base-multilingual-cased}, XLM-R\footnote{https://huggingface.co/FacebookAI/xlm-roberta-base}, and mT5 models published in HuggingFace to obtain the transformer-based baseline models. Especially, mT5 is available in various sizes. We used mT5-large\footnote{https://huggingface.co/google/mt5-large} as this model was the largest model among the variations which we could accommodate with the same resources utilized for training and testing other models. Similarly, the open-source large language models (LLMs) available in HuggingFace were used to obtain the LLM baselines, Llama2\footnote{https://huggingface.co/meta-llama/Llama-2-7b-chat-hf}, Mistral 7B\footnote{https://huggingface.co/mistralai/Mistral-7B-Instruct-v0.2}, and Phi3\footnote{https://huggingface.co/microsoft/Phi-3-mini-4k-instruct}. We used the same experimental setting of \cite{li2024factfinders}. The prompt \ref{prompt} used to fine-tune LLMs is a modified version of \cite{li2024factfinders}.

\begin{prompt}[H]
\begin{footnotesize}
\begin{verbatim}
### Instruction:
Evaluate whether the input text contains 
information or claims that can be verified 
through fact-checking. If the statement presents 
assertions, facts, or claims, respond with 'Yes'. 
If the statement is purely opinion-based, trivial,
or does not contain any verifiable information or 
claims, respond with 'No'.

### Input Sentence: <input sentence>

### Response: <Yes/No>
\end{verbatim}
  \caption{Verifiable Claim Detection}
  \label{prompt}
\end{footnotesize}
\end{prompt}

All the models were fine-tuned by extracting the vector representation of the special token from the final layer and applying a fully connected layer on it for classification. We used only the encoder of the mT5 model for obtaining vector representation of input words. For Named Entity Recognition (NER), we used the resources published for MultiNERD\footnote{https://github.com/Babelscape/multinerd/tree/master} for fine-tuning the fine-grained NER model. Similarly, the mGENRE\footnote{https://huggingface.co/facebook/mgenre-wiki} model published in HuggingFace was used for Entity Linking (EL).

\subsection{Model Comparison}\label{sec:model_comaprison}

Table \ref{tab:efficiency_comparison} compares claim detection models based on the number of parameters, multilingual support, and training time. Both \textit{EXN-Claim} and EXP-Claim incur an additional average latency of 5.2 ms per sentence for named entity recognition. Furthermore, \textit{EXP-Claim} requires an average of 0.15 seconds per entity to perform entity linking.

\subsection{Hyperparameters}\label{subsec:environment_setting}

\begin{table}[H]
\centering
\footnotesize
\begin{tabular}{lll}\hline
Hyperparameter &  Value \\ \hline
Projection Size          & 256           \\
Learning Rate    & 3e-5          \\
Entity Embedding Size & 128 (\textit{X, EXN}), 256 (\textit{EXP})\\
Entity Linking Threshold & -0.15 \\ \hline
\end{tabular}
\caption{ Hyperparameters Used}\label{tab:hyperparamters}
\end{table}

Table \ref{tab:hyperparamters} lists the hyperparameters used. Following \cite{bert}, we set the maximum length of the input sequence to 128, and use Adam as the optimizer. We train the models for 30 epochs and choose the model with the lowest validation loss as the best model. 

\end{document}